\pdfoutput=1

\documentclass[11pt]{article}

\usepackage[final]{acl}

\usepackage{times}
\usepackage{latexsym}

\usepackage[T1]{fontenc}

\usepackage[utf8]{inputenc}

\usepackage{microtype}

\usepackage{inconsolata}

\usepackage{graphicx}

\usepackage{booktabs}       
\usepackage{multirow}
\usepackage{wrapfig}

\definecolor{amethyst}{rgb}{0.6, 0.4, 0.8}

\definecolor{cvprblue}{rgb}{0.21,0.49,0.74}
\definecolor{weakorange}{RGB}{255,230,200}
\definecolor{weakgray}{RGB}{240,240,240} 
\definecolor{sacramentostategreen}{rgb}{0.0, 0.34, 0.25}
\definecolor{navyblue}{rgb}{0.0, 0.0, 0.5}
\definecolor{colbest}{rgb}{0.1, 0.6, 0.1}
\definecolor{colworst}{rgb}{0.75, 0, 0}
\definecolor{oursrow}{HTML}{EBF5F8}
\definecolor{palegreen}{rgb}{0.6, 0.98, 0.6}
\definecolor{calmpalegreen}{rgb}{0.8, 0.95, 0.8}
\definecolor{sacramentostategreen}{rgb}{0.0, 0.34, 0.25}
\definecolor{navyblue}{rgb}{0.0, 0.0, 0.5}
\definecolor{amaranth}{rgb}{0.9, 0.17, 0.31}
\definecolor{amethyst}{rgb}{0.6, 0.4, 0.8}
\definecolor{ao(english)}{rgb}{0.0, 0.5, 0.0}
\definecolor{blue(ryb)}{rgb}{0.01, 0.28, 1.0}
\definecolor{cardinal}{rgb}{0.77, 0.12, 0.23}
\definecolor{cobalt}{rgb}{0.0, 0.28, 0.67}
\definecolor{mediumblue}{rgb}{0.0, 0.0, 0.8}

\hypersetup{
colorlinks=true,
linkcolor=navyblue,
citecolor=navyblue,
filecolor=navyblue,
urlcolor=navyblue}

\usepackage{amsfonts}
\usepackage{amsmath}
\usepackage{enumitem}
\usepackage{tabularx}
\usepackage{tcolorbox}
\tcbuselibrary{skins}

\newtcolorbox{insightbox}{
    enhanced,
    colback=white, 
    colframe=black, 
    arc=0mm, 
    fonttitle=\bfseries, 
    coltitle=black, 
    attach title to upper, 
    title={Insight. }, 
    sharp corners=all, 
    top=0mm, 
    bottom=0mm, 
    left=0mm,
    right=0mm, 
}

\usepackage{soul}

\usepackage{tikz}
\usepackage{amsmath}

\DeclareMathOperator*{\argmin}{arg\,min}
\usepackage{cleveref}
\usepackage{bbold} 

\definecolor{redhighlight}{HTML}{FFBFBF}
\definecolor{bluehighlight}{HTML}{CCFFFF}
\definecolor{newblue}{HTML}{CDF0F1}  
\definecolor{newred}{HTML}{FAE4E4}  

\newcommand{\hlred}[1]{\sethlcolor{redhighlight}\hl{#1}}
\newcommand{\hlblue}[1]{\sethlcolor{bluehighlight}\hl{#1}}
\newcommand{\hlnewred}[1]{\sethlcolor{newred}\hl{#1}}
\newcommand{\hlnewblue}[1]{\sethlcolor{newblue}\hl{#1}}

\title{Resampled Datasets Are Not Enough:\\Mitigating Societal Bias Beyond Single Attributes}

\author{%
\textbf{Yusuke Hirota$^{1*}$}
\quad
\textbf{Jerone T. A. Andrews$^{2}$}
\quad
\textbf{Dora Zhao$^{3*}$}
\quad
\textbf{Orestis Papakyriakopoulos$^{4*}$} 
\\
\textbf{Apostolos Modas$^{2}$}
\quad
\textbf{Yuta Nakashima$^{1}$}
\quad
\textbf{Alice Xiang$^{2}$} 
\\
\\
$^{1}$Osaka University\quad $^{2}$Sony AI\quad $^{3}$Stanford University \quad $^{4}$Technical University of Munich
}

\begin{document}
\maketitle

\renewcommand{\thefootnote}{\fnsymbol{footnote}}
\footnotetext[1]{Work done at Sony AI. Correspondence to Yusuke Hirota: \texttt{y-hirota@is.ids.osaka-u.ac.jp}.}  

\renewcommand{\thefootnote}{\arabic{footnote}}

\begin{abstract}
    We tackle societal bias in image-text datasets by removing spurious correlations between protected groups and image attributes. Traditional methods only target labeled attributes, ignoring biases from unlabeled ones. Using text-guided inpainting models, our approach ensures protected group independence from all attributes and mitigates inpainting biases through data filtering.  Evaluations on multi-label image classification and image captioning tasks show our method effectively reduces bias without compromising performance across various models.
\end{abstract}

\section{Introduction}
\label{sec:intro}


Models trained on biased data can develop prediction rules based on spurious correlations (i.e., associations devoid of causal relationships), perpetuating and amplifying harmful stereotypes~\citep{zhao2017mals}. For example, image captioning models may generate gendered captions by associating gender with depicted activities~\citep{zhao2023men}, location~\citep{hendricks2018women}, or objects~\citep{wang2021biasamp}. 
Dataset-level bias mitigation aims to reduce spurious correlations between labeled image attributes (e.g., \texttt{teddy bear}) and protected groups (e.g., \texttt{woman}). Resampling approaches balance the co-occurrence of each attribute with each  group~\citep{agarwal2022does,wang2020towards}. However, models can still exploit correlations between groups and sets of attributes (e.g., \texttt{man} with $\{\texttt{dog}, \texttt{pizza}, \texttt{couch}\}$), even when individual attributes are balanced~\citep{zhao2023men}. Moreover, spurious correlations extend to unlabeled attributes, which current strategies do not address---e.g., gender disparities in image color statistics~\citep{meister2023gender} or the person-to-object spatial distances~\citep{wang2020revise}.

While equal group distributions in real-world datasets are challenging to achieve, generative text-to-image models now enable targeted image modifications~\cite{rombach2022high,brooks2023instructpix2pix,couairon2022diffedit}. For example, bias detection methods alter image subjects' appearance to assess counterfactual fairness~\cite{joo2020gender} or model bias~\citep{smith2023balancing,brinkmann2023multidimensional}. However, manipulating individuals' appearances without consent raises significant ethical and privacy concerns~\citep{andrews2023ethical,yew2022regulating,sobel2020taxonomy,ramaswamy2020debiasing,orekondy2018connecting,oh2016faceless}.

 \begin{figure*}[t]
   \centering
   \includegraphics[clip, width=1.0\textwidth]{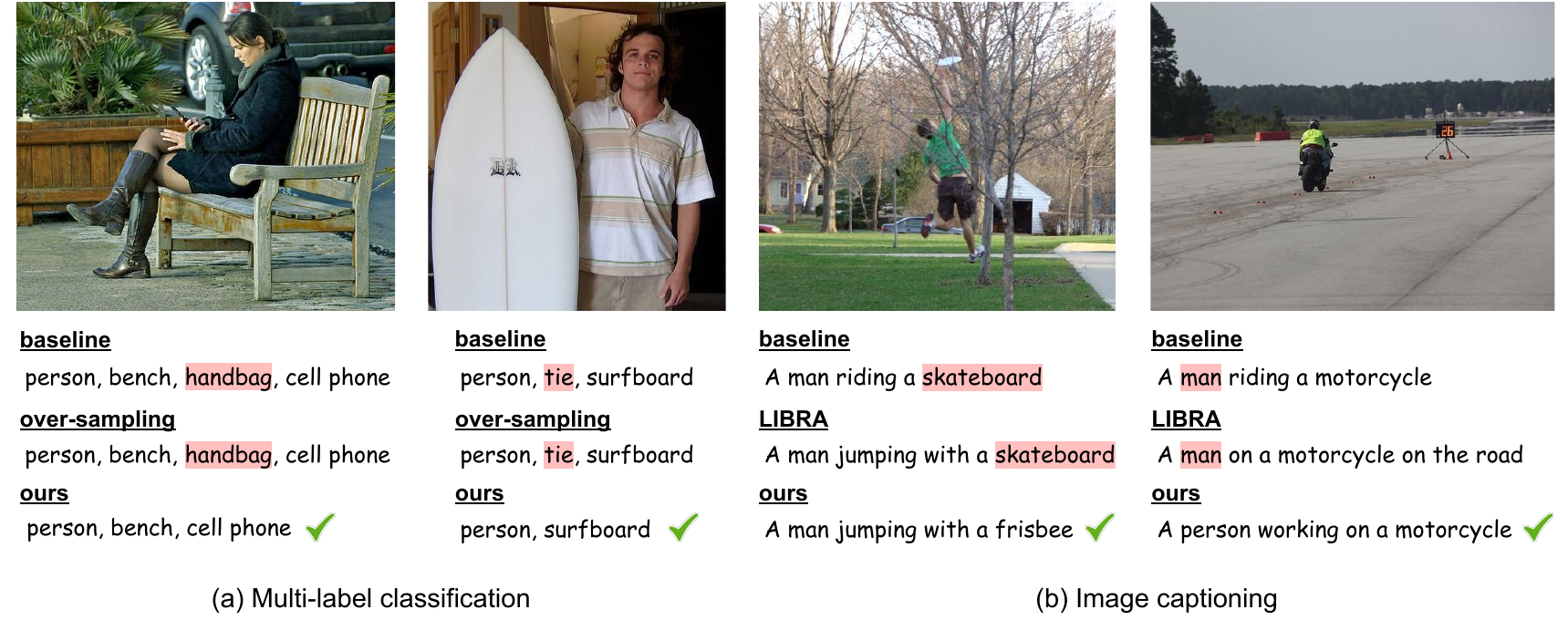}
   \caption{(a) Predicted objects by baseline ResNet-50 and with bias mitigation, i.e., over-sampling~\cite{wang2020towards} versus our method. (b) Generated captions by baseline ClipCap and with bias mitigation, i.e., LIBRA~\cite{hirota2023model} versus our method. Incorrect predictions, possibly affected by gender-object correlations, are in \hlred{red}.}
   \label{fig:fig1-quali}
   \vspace{-5pt}
 \end{figure*}

To address these challenges, we create training datasets with text-guided inpainting~\citep{rombach2022high}, ensuring attribute distributions are independent of protected groups. Using masked person images and text prompts, we generate counterfactual images by inpainting only the masked regions, addressing ethical concerns of altering nonconsensual persons and ensuring equal representation of protected groups across attributes. We introduce data filters to mitigate biases from generative text-guided inpainting models~\cite{bianchi2023easily,cho2023dall,bansal2022well,luccioni2023stable}, evaluating images based on adherence to prompts, preservation of attributes and semantics, and color fidelity, validated by human evaluators. Unlike prior work~\citep{wang2019balanced,wang2020towards,zhao2023men,agarwal2022does}, training on our counterfactual data decorrelates both labeled and unlabeled attributes from protected groups without impacting model performance. Comprehensive evaluations show our approach significantly reduces prediction rules based on spurious correlations in multi-label classification and image captioning across various architectures (e.g., ResNet-50~\cite{resnet}, Swin Transformer~\cite{liu2021swin}), datasets (COCO~\citep{lin2014microsoft}, OpenImages~\cite{krasin2017openimages}), and protected groups (gender, skin tone). Our key contributions are summarized as follows:
\begin{itemize}[label=\textbullet, left=0pt, labelsep=10pt, itemsep=0pt, parsep=0pt, partopsep=0pt, align=left, topsep=0pt]
\item Introducing a framework for generating synthetic training datasets with group-independent image attribute distributions.
\item Proposing data filtering to mitigate biases introduced by generative inpainting models.
\item Conducting quantitative experiments, demonstrating significant bias reduction in classification and captioning tasks compared to baselines.
\item Identifying limitations of training on combined real and synthetic datasets, emphasizing the need for cautious synthetic data augmentation.
\end{itemize}

\subsection{Related Work}
\label{sec:related}

Societal bias in datasets, characterized by demographic imbalances leading to spurious correlations, has been extensively studied~\cite{devries2019everyone,birhane2024into, prabhu2020large,birhane2021multimodal,wang2020revise,meister2023gender}. These biases persist and can be exacerbated by multi-label classifiers~\cite{zhao2017mals,de2019does,wang2019balanced} and image captioning models~\cite{zhao2021captionbias,hendricks2018women,hirota2022quantifying},
disproportionately impacting historically marginalized groups such as women and individuals with darker skin tones~\cite{garcia2023uncurated,ross2020measuring}.

Two common approaches to bias mitigation are dataset-level and model-level. Dataset-level approaches leverage generative adversarial networks (GANs), counterfactual training dataset augmentation, and resampling. GANs create synthetic images to balance datasets and mitigate spurious correlations~\cite{ramaswamy2021fair,sattigeri2019fairness,sharmanska2020contrastive}, counterfactual data augmentation generates alternative scenarios to address biases~\cite{kaushik2019learning,wang2021robustness}, and resampling balances the co-occurrence of attributes and protected groups~\cite{agarwal2022does,wang2020towards}. Model-level approaches reduce bias through corpus-level constraints~\cite{zhao2017mals}, adversarial debiasing~\cite{wang2019balanced,hendricks2018women,tang2021mitigating,alvi2018turning}, domain discriminative/independent training~\cite{wang2020towards}, modified loss functions~\cite{lin2017focal,cui2019class,sagawa2019distributionally}, and model output editing~\cite{hirota2023model}. 
However, despite these advancements, existing mitigation methods focus on single labeled attributes, which can inadvertently increase models’ reliance on spurious correlations between protected groups and combinations of attributes~\cite{zhao2023men} or unlabeled attributes~\cite{meister2023gender}. 

Recent progress in text-to-image generative models has enabled targeted image manipulation~\cite{rombach2022high,brooks2023instructpix2pix,couairon2022diffedit}, which can help address bias in multi-modal datasets. Nonetheless, these models have also been shown to perpetuate harmful stereotypes~\cite{mandal2023multimodal,zhang2023auditing,wang2023t2iat,struppek2022biased,ungless2023stereotypes,naik2023social,seshadri2023bias,friedrich2023fair}. 
In contrast to prior bias mitigation work, we use text-guided inpainting to generate synthetic training datasets that ensure equal representation of protected groups across all attribute combinations, whether labeled or unlabeled. To mitigate inpainting biases, we propose data filters, producing higher quality and less biased synthetic data. We go beyond previous work focused solely on gender bias mitigation~\cite{joo2020gender,smith2023balancing,brinkmann2023multidimensional} by also addressing skin tone biases.

\section{Method}
\label{sec:method}

\begin{figure*}[t]
  \centering
  \includegraphics[clip, width=0.95\textwidth]{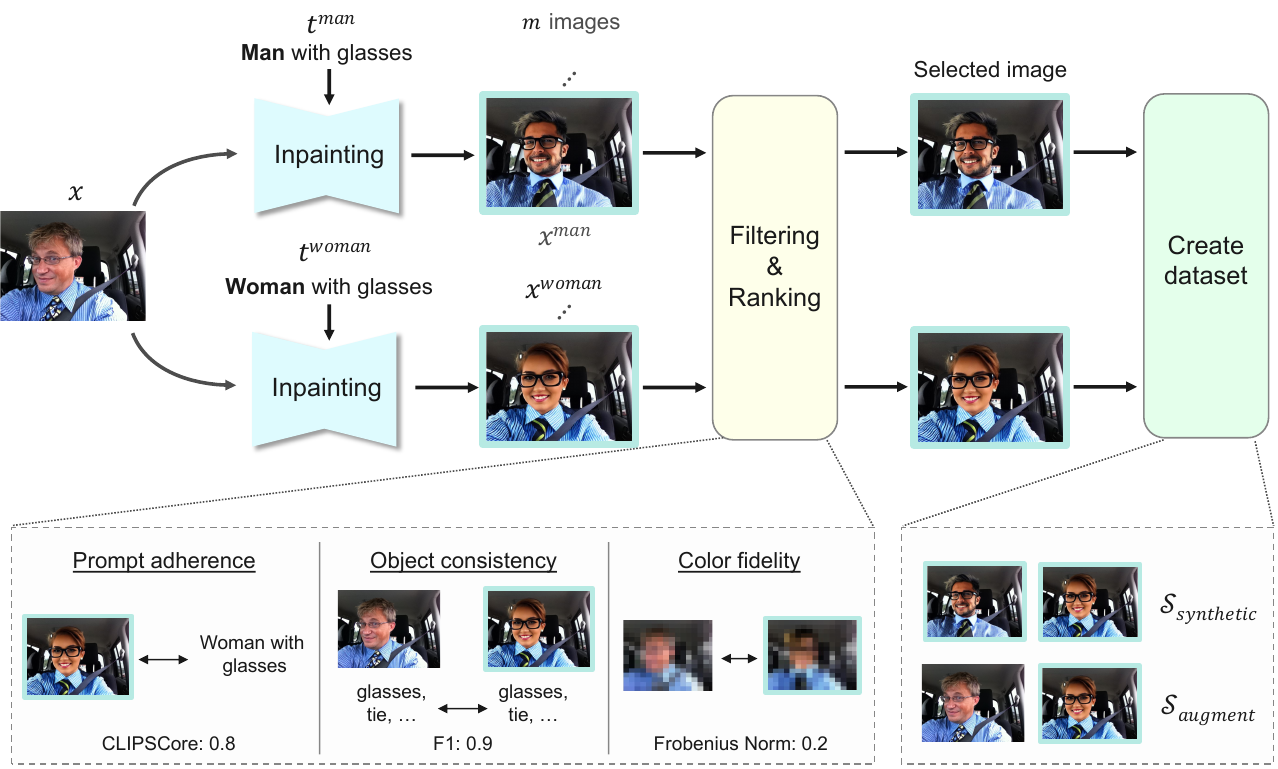}
  \vspace{-2pt}
  \caption{Overview of our pipeline for binary gender as a protected attribute. Original images are inpainted to synthesize diverse groups, maintaining consistent context. Synthesized images (highlighted in \hlblue{blue}) are ranked using filters to select high-quality, unbiased samples (Module: Filtering \& Ranking). Selected images are then used to construct datasets with group-independent image attribute distributions (Module: Create dataset).}
  \label{fig:method}
   \vspace{-5pt}
\end{figure*}

We create training datasets with group-independent image attribute distributions by using masked person images and text prompts with an off-the-shelf diffusion model, as outlined in~\Cref{fig:method}. 

\subsection{Resampled Datasets Are Not Enough} 
\label{sec:preliminaries}
We denote an image by $x \in \mathcal{X}$, a protected group by $g \in \mathcal{G}$, and an image attribute by $a\in\mathcal{A}$. A spurious correlation exists if $p_\mathcal{X}(a \mid g) \neq p_\mathcal{X}(a)$, indicating biases in the data. 
Resampling aims to remove these biases by adjusting the sampling process so that $p_\mathcal{X}(a \mid g) = p_\mathcal{X}(a)$ for all $g$~\cite{de2019does,wang2020towards}. 
This is done using a limited set of labeled attributes $\mathcal{O} \subset \mathcal{A}$, where attributes $a$ are drawn from a distribution $q(a)$ over $\mathcal{O}$ and groups $g$ are drawn from a uniform distribution $u(g)$ over $\mathcal{G}$ such that $\mathcal{X}^\prime = \{x \sim p_\mathcal{X}(x \mid g, a) \mid a \sim q(a), g \sim u(g)\}$.
This ensures $p_{\mathcal{X}^\prime}(a \mid g) = q(a)$ for $a \in \mathcal{O}$ and $g \in \mathcal{G}$. 
However, this method has a limitation: it does not account for $a$ being an unlabeled attribute or a combination of labeled and unlabeled attributes, making it difficult to sample $x$ from $p_\mathcal{X}(x \mid g,a)$ due to insufficient information about $a$. 
In short, while resampling can reduce biases, it is not always enough, especially when dealing with unlabeled or mixed attributes.

\subsection{Text-Guided Inpainting}
\label{sec:method-1}

Suppose $\mathcal{D} = \{(x_i, \omega_i, a_i, t_i^{(g)}) \mid 1\leq i\leq n \}$ is a training set, where $x\in\mathbb{R}^d$ is an image, $\omega\in [0,1]^d$ is a person mask, $a$ is a labeled image attribute, a combination of labeled attributes, or an unlabeled attribute, and $t^{(g)}$ is a text prompt containing a protected group-specific word $g$. To create a dataset with group-independent image attribute distributions, we utilize a text-guided inpainting model~\cite{rombach2022high}. This model, guided by $t^{(g)}$, inpaints $\omega$ in $x$ with a synthetic person from protected group $g$ described in $t^{(g)}$. For each tuple in $\mathcal{D}$, we generate $m \in \mathbb{N}^+$ versions for each $g \in \mathcal{G}$, resulting in $m \cdot |\mathcal{G}|$ samples:
\begin{equation} \label{eq:counterfactual-dataset}
    \begin{split}
        \mathcal{D}_{\text{synthetic}} &=  \{ (x_i^{(j,g^\prime)}, \omega_i, a_i, t_i^{(g^\prime)}) \\ 
        &\mid 1\leq i\leq n, g^\prime\in\mathcal{G}, 1\leq j \leq m\},
    \end{split}
\end{equation}
where $x_i^{(j,g^\prime)}$ denotes the $j$-th inpainted version of $x_i\in\mathcal{X}$ for $g^\prime$ and $t_i^{(g^\prime)}$ the modified text prompt where $g$ in $t_i^{(g)}$ is replaced with $g^\prime$.

\subsection{Societal Bias Data Filtering}
\label{sec:method-2}

Text-to-image generative models often perpetuate societal biases, portraying certain groups stereotypically, such as depicting women in brighter clothing~\cite{bianchi2023easily,cho2023dall,bansal2022well,luccioni2023stable}. Since these biases remain largely unaddressed~\cite{smith2023balancing,brinkmann2023multidimensional}, we set $m>1$ in \Cref{eq:counterfactual-dataset} to generate multiple variations for each group. We propose filters to select the least biased inpainted images, evaluating images based on adherence to text prompts, preservation of attributes and semantics, and color fidelity. Specifically, for each tuple $(i,g^\prime)$, we select the highest quality and least biased version among the $m$ versions to create a training dataset:
\begin{equation} \label{eq:pruned-counterfactual-dataset}
\begin{split} 
    \mathcal{S}_{\text{synthetic}} &=  \{ (x_i^{(j^{\star}, g^\prime)}, \omega_i, a_i, t_i^{(g^\prime)})  \in \mathcal{D}_{\text{synthetic}} \\
    &\mid \forall (i,g^\prime), j^\star \},
\end{split}
\end{equation}
where $j^\star = \argmin_j \sum_k c_k \cdot r(s_k^{(i,j,g^\prime)})$, $c_k \in \mathbb{R}$ are weights assigned to filters $s_k$, $s_k^{(i,j,g^\prime)}$ is the score obtained from applying filter $s_k$ to image $x_i^{(j,g^\prime)}$ for group $g^\prime$, and $r(s_k^{(i,j,g^\prime)})$  is the rank of the score for $(i, g^\prime)$ in descending order, with lower ranks indicating less bias. Here, $x_i^{(j^\star, g^\prime)}$ is the selected inpainted image for tuple $(i, g^\prime)$ that minimizes the sum of the ranks of the weighted filter scores, with $j^\star$ representing the index of the selected candidate image for tuple $(i, g^\prime)$. 

Rather than creating an entire dataset of synthetic samples, we can augment $\mathcal{D}$:
\begin{equation} \label{eq:pruned-counterfactual-dataset-augment}
\begin{split} 
    \mathcal{S}_{\text{augment}} &=  \mathcal{D} \cup \{ (x_i^{(j^{\star}, g^\prime)}, \omega_i, a_i, t_i^{(g^\prime)})  \in \mathcal{D}_{\text{synthetic}} \\
    &\mid \forall (i,g^\prime\neq g), j^\star \}.
\end{split}
\end{equation}
The condition  $g^\prime \neq g$  ensures that we only add inpainted images to $\mathcal{D}$ for groups different from those originally present in $x_i$.
In contrast to resampling, $\mathcal{S}_{\text{synthetic}}$ and $\mathcal{S}_{\text{augment}}$ ensure $p_{\mathcal{X}'}(a \mid g) = p_{\mathcal{X}}(a)$ for all $g \in \mathcal{G}$ without making assumptions about $\mathcal{A}$. 
Our proposed filters are introduced below.

\paragraph{Prompt Adherence.}
To evaluate the semantic alignment between $x_i^{(j, g^\prime)}$ and $t_i^{(g^\prime)}$, we use CLIPScore~\cite{hessel2021clipscore}, which computes the cosine similarity between their CLIP embeddings~\cite{radford2021learning}. Formally,
\begin{equation}
    s_{\text{prompt}}^{(i,j,g^\prime)} = \phi(x_i^{(j, g^\prime)}) \cdot \psi(t_i^{(g^\prime)}) \in [-1,1],
\end{equation}
where $\phi$ and $\psi$ are CLIP's vision and text encoders, respectively. If $s_{\text{prompt}}^{(i,j,g^\prime)} > s_{\text{prompt}}^{(i,j^\prime,g^\prime)}$, then $x_i^{(j, g^\prime)}$ better reflects the content described in $t_i^{(g^\prime)}$.

\paragraph{Object Consistency.}
To prevent the introduction of spurious correlations, such as generating objects not mentioned in $t_i^{(g^\prime)}$ or reinforcing stereotypes~\cite{bianchi2023easily,cho2023dall,bansal2022well}, we assess the object similarity between predicted objects in $x_i^{(j, g^\prime)}$ and $x_i$. Concretely, we compute the F1 score~\cite{sokolova2006beyond} using a pretrained object detector~\cite{zhou2022detecting}, denoted $\eta$:
\begin{equation}
    s_{\text{object}}^{(i,j,g^\prime)} = \text{F1}[\eta(x_i^{(j, g^\prime)}), \eta(x_i)] \in [0,1].
\end{equation}
If $s_{\text{object}}^{(i,j,g^\prime)} > s_{\text{object}}^{(i,j^\prime,g^\prime)}$, then $x_i^{(j, g^\prime)}$ better preserves the integrity of the original unmasked scene in $x_i$. 

\paragraph{Color Fidelity.}
Generative models can introduce subtler biases~\cite{bansal2022well,bianchi2023easily}, including those related to color~\cite{meister2023gender}. Addressing color biases is crucial as color choices can implicitly carry cultural or gendered connotations. To mitigate this, we downsample $x_i^{(j, g^\prime)}$ and $x_i$ to $14 \times 14$ pixels to focus on color rather than fine details, then measure the color difference using the Frobenius norm:
\begin{equation}
    s_{\text{color}}^{(i,j,g^\prime)} = \|(x_i^{(j, g^\prime)})_{\downarrow 14 \times 14} - (x_i)_{\downarrow 14 \times 14}\|_\text{F}^{-1}.
\end{equation}
If $s_{\text{color}}^{(i,j,g^\prime)} > s_{\text{color}}^{(i,j^\prime,g^\prime)}$, then $x_i^{(j, g^\prime)}$ has better color fidelity to the original unmasked scene in $x_i$.

\begin{table*}[t]
\footnotesize
\centering
\begin{tabularx}{0.89\textwidth}{l r r r r r r r r r r r}
\toprule
& \multicolumn{3}{c}{ResNet-50} && \multicolumn{3}{c}{Swin-T} && \multicolumn{3}{c}{ConvNeXt-B} \\ 
 \cline{2-4} 
 \cline{6-8}
 \cline{10-12}
& \multirow{1.3}{*}{mAP} & \multirow{1.3}{*}{Ratio} & \multirow{1.3}{*}{Leakage} && \multirow{1.3}{*}{mAP} & \multirow{1.3}{*}{Ratio} & \multirow{1.3}{*}{Leakage}  && \multirow{1.3}{*}{mAP} & \multirow{1.3}{*}{Ratio} & \multirow{1.3}{*}{Leakage} \\
\midrule
Original  & \underline{66.4}  & 6.3  & 13.4  && \textbf{72.8} & 4.0  & 14.3 && \textbf{76.3} & 4.6 & 18.2 \\
\midrule
Adversarial & 63.3 & — & \textbf{3.3}  && 67.8  & --- & \textbf{4.4} && 69.6 & --- & \textbf{4.7} \\
DomDisc & 57.4 & 4.1 & 15.4 && 65.4 & 4.6 & 16.8 && 68.8 & 4.5 & 19.1 \\
DomInd  & 60.4 & 2.8 & 10.4 && 67.9 & 3.8 & 11.4 && 72.6 & 5.9 & 15.0 \\
Upweight & 64.9 & 9.1 & 8.3 && 71.5 & 6.3 & 9.8 && 75.0 & 5.6 & 12.9 \\
Focal  & 66.1 & 6.3 & 12.0 && \underline{72.2} & 3.8 & 13.3 && \underline{76.2} & 3.8 & 16.2 \\
CB  & 63.0 & 4.3 & 10.9 && 69.6 & 3.5 & 12.3 && 73.8 & 3.5 & 14.7 \\
GroupDRO  & 64.1 & 3.0 & 11.4 && 70.8 & \underline{1.5} & 12.6 && 75.3 & 4.2 & 16.4 \\
\midrule
Over-sampling & 62.6 & 3.8 & 9.7 && 69.9 & 2.6 & 10.5 && 73.5 & 3.4 & 13.7 \\
Sub-sampling & 58.3 & \underline{2.0} & 12.2 && 64.4 & 1.8 & 11.6 && 66.3 & \underline{2.2} & 18.2 \\
$\mathcal{S}_{\text{augment}}$ (Ours) & \textbf{66.9} & 4.6 & 8.1 && \textbf{72.8} & 3.1 & 10.5 && \textbf{76.3} & \underline{2.2} & 11.3 \\
$\mathcal{S}_{\text{synthetic}}$ (Ours) & 66.0 & \textbf{1.1} & \underline{7.5} && 71.9 & \textbf{1.4} & \underline{8.4} && 75.5 & \textbf{1.2} & \underline{8.2} \\
\bottomrule
\end{tabularx}
\vspace{-4pt}
\caption{Classification performance and gender bias scores of ResNet-50, Swin-T, and ConvNeXt-B backbones on COCO. Ratio is inapplicable to Adversarial due to its gender prediction module for mitigation. \textbf{Bold} and \underline{underline} represent the best and second-best, respectively. For an unbiased model, $\text{Ratio} = \text{1}$ and $\text{Leakage} = \text{0}$.}
\label{tab:new-mlc-coco-main}
\end{table*}

\begin{table*}[t]
\centering
\footnotesize
\begin{tabularx}{0.95\textwidth}{l r r r r r r r r r r r r r r}
\toprule
& \multicolumn{4}{c}{ClipCap} && \multicolumn{4}{c}{BLIP-2} && \multicolumn{4}{c}{Transformer} \\ 
\cline{2-5}
\cline{7-10}
\cline{12-15}
& \multirow{1.3}{*}{M} & \multirow{1.3}{*}{CS}  & \multirow{1.3}{*}{Ratio} & \multirow{1.3}{*}{LIC} && \multirow{1.3}{*}{M} & \multirow{1.3}{*}{CS} & \multirow{1.3}{*}{Ratio} & \multirow{1.3}{*}{LIC} && \multirow{1.3}{*}{M} & \multirow{1.3}{*}{CS}  & \multirow{1.3}{*}{Ratio} & \multirow{1.3}{*}{LIC} \\
\midrule
Original  & \textbf{29.1} & \underline{75.1} & 2.5 & 2.2 && \textbf{29.5} & 75.1 & 5.7 & 4.7 && \underline{26.9} & \underline{71.5} & 4.7 & 4.7  \\
\midrule
LIBRA & 28.9 & 74.9 & 6.5 & \underline{0.5} && 29.0 & \textbf{75.4} & 6.3 & \textbf{1.9} && \textbf{27.4} & \textbf{73.4} & 6.7 & 2.3 \\
\midrule
Over-sampling   & 28.6 & 74.7 & 3.2 & 3.5 && 28.7 & 74.1 & 3.8 & 3.0 && 26.2 & 70.6 & 4.1 & 1.6 \\
Sub-sampling & 28.0 & 74.0 & \underline{1.4} & 4.1 && 28.3 & 74.5 & \underline{1.4} & 3.2 && 25.0 & 69.7 & \underline{2.0} & 3.9 \\
$\mathcal{S}_{\text{augment}}$ (Ours) & \underline{29.0} & 75.0 & 2.5 & 1.7 && \underline{29.4} & \underline{75.3} & 2.9 & 3.8 && 26.2 & 71.1 & 2.6 & \underline{1.5} \\
$\mathcal{S}_{\text{synthetic}}$ (Ours) & 28.5 & \textbf{75.3} & \textbf{1.3} & \textbf{0.3} && 29.3 & 75.0 & \textbf{1.2} & \underline{2.5} && 25.7 & 70.9 & \textbf{1.4} & \textbf{0.5} \\
\bottomrule
\end{tabularx}
\vspace{-4pt}
\caption{Captioning quality and gender bias scores of ClipCap, BLIP-2, and Transformer backbones on COCO. M and CS denote METEOR and CLIPScore. \textbf{Bold} and \underline{underline} represent the best and second-best, respectively. For an unbiased model, $\text{Ratio} = \text{1}$ and $\text{LIC} = \text{0}$.}
\label{tab:new-ic-coco-main}
\end{table*}

\section{Experiments}
\label{sec:experiments}
Building on prior research~\cite{zhao2017mals,wang2019balanced,zhao2023men,hendricks2018women,zhao2021captionbias,tang2021mitigating}, we evaluate our synthetic dataset creation method on multi-label image classification and image captioning tasks using quantitative metrics, human studies, qualitative comparisons, and effectiveness analysis. Evaluations are conducted on test sets of real data.

\paragraph{Implementation Details.}
We inpaint the largest person in the image based on bounding box size, and if the second largest person exceeds 55,000 pixels, we also inpaint that region, using the \texttt{person} label for COCO. For image generation, we create $m=\text{30}$ inpainted images per group (e.g., {\texttt{woman}, \texttt{man}}) using guidance scales of 7.5, 9.5, and 15.0 to ensure diversity. Filter weights are set to 1 (i.e., $c_k=1$ for all $k$), contributing equally. Results are based on five models trained with different random seeds. More details are in \Cref{sec:app-detail,sec:app-exp}.

\subsection{Multi-Label Classification}
\label{sec:mlc}
\paragraph{Experimental Setup.}
We focus on gender bias using the COCO dataset, retaining only images with gender-specific terms (e.g., \texttt{woman}, \texttt{man}) in their captions. This results in 28,487/13,487 train/test samples. We focus on objects co-occurring with these terms, yielding 51 objects. ResNet50, Swin Transformer Tiny (Swin-T), and ConvNext models are fine-tuned using early stopping. Performance is assessed using mean average precision (mAP).
Bias is quantified using leakage and ratio. Leakage measures how much the model's predictions amplify the group's information compared to the ground truth. A gender classifier $f_g(y)$, predicting gender group $g$ from input $y$ (i.e., set of objects), is trained on a training set $\mathcal{T} = \{(y, g)\}$. For the test set $\mathcal{T}'$, the model's leakage score is:
\begin{equation}\label{eq:leak}
\begin{aligned}
\text{LK}_\text{M} &= \frac{1}{|\mathcal{T}'|} \sum_{(y, g) \in \mathcal{T}'}\!\!\!\!f_g(y) \mathbb{1}\left[\arg\max_{g'} f_{g'}(y) = g\right]
\end{aligned}
\end{equation}
The leakage score for the original dataset, $\text{LK}_\text{D}$, is similarly computed. The final leakage is $\text{Leakage} = \text{LK}_\text{M} - \text{LK}_\text{D}$.
Higher leakage indicates greater model exploitation of protected group information. Ratio measures the exploitation of attribute information for group prediction. By masking individuals in test images and measuring the bias in group predictions (e.g., \#\texttt{man}-to-\#\texttt{woman} ratio), deviations from a ratio of 1 indicate attribute exploitation. We report $\text{Ratio}=\max(r,r^{-1})$, where $r$ is the observed ratio. This captures the magnitude of deviation from unbiased predictions consistently. 

We compare our method with existing bias mitigation techniques, including dataset-level methods (Over-sampling~\cite{wang2020towards}, Sub-sampling~\cite{agarwal2022does}) and model-level methods such as adversarial debiasing~\cite{wang2019balanced} (Adversarial), domain-independent training~\cite{wang2020towards} (DomInd), domain discriminative training~\cite{wang2020towards} (DomDisc), loss upweighting~\cite{byrd2019effect} (Upweight), focal loss~\cite{lin2017focal} (Focal), class-balanced loss~\cite{cui2019class} (CB), and group DRO~\cite{sagawa2019distributionally} (GroupDRO). Additional results on the OpenImages dataset and skin tone bias mitigation are provided in \Cref{sec:app-mlc}, demonstrating consistent conclusions.

\paragraph{Results.}
Results are shown in~\Cref{tab:new-mlc-coco-main}. Our method, $\mathcal{S}_{\text{synthetic}}$, achieves the best balance by significantly improving both ratio and leakage while maintaining a high mAP.  Specifically, $\mathcal{S}_{\text{synthetic}}$ achieves a near-ideal ratio of 1.1, low leakage of 7.5, and an mAP of 66.0 for ResNet-50, with similar trends observed for Swin-T and ConvNeXt-B.

Adversarial debiasing achieves lower leakage scores by removing gender information from intermediate representations. However, this method reduces mAP, indicating that object information may also be inadvertently removed. Over-sampling and sub-sampling methods address class imbalance but at the cost of model performance. Sub-sampling, in particular, reduces the ratio compared to over-sampling but results in worse mAP and increased leakage. This is likely due to the loss of diversity and information in the training data, which forces the model to rely more on the remaining features, increasing the influence of protected attributes.

In contrast, $\mathcal{S}_{\text{synthetic}}$ generates diverse, high-quality synthetic samples, effectively balancing bias and variance. This approach avoids the pitfalls of other methods, resulting in superior performance metrics. While $\mathcal{S}_{\text{augment}}$ performs similarly to the original dataset, it performs worse in terms of ratio and leakage compared to $\mathcal{S}_{\text{synthetic}}$.

\subsection{Image Captioning}
\label{sec:ic}

\paragraph{Experimental Setup.}
Using the COCO dataset (\Cref{sec:mlc}), we benchmark captioning models ClipCap, BLIP-2, and Transformer, which are fine-tuned using early stopping. Performance is evaluated with METEOR and CLIPScore. Bias is quantified using LIC and ratio, where LIC is a leakage-based metric that assesses the generation of group-stereotypical captions compared to ground-truth captions (i.e., $y$ is a caption in \Cref{eq:leak}), and predicted group-related terms (e.g., \texttt{woman}) in captions used to compute ratio.

Bias mitigation baselines include dataset-level methods (Over-sampling, Sub-sampling) and the current state-of-the-art model-level method LIBRA~\cite{hirota2023model}. LIBRA is a model-agnostic debiasing framework designed to mitigate bias amplification in image captioning by synthesizing gender-biased captions and training a debiasing caption generator to recover the original captions. Detailed results for skin tone bias mitigation, along with fine-tuning specifics, are provided in \Cref{sec:app-ic}, showcasing the generalizability of our approach.

\paragraph{Results.}
Results are shown in \Cref{tab:new-ic-coco-main}. Our method, $\mathcal{S}_{\text{synthetic}}$, significantly improves both ratio and LIC while maintaining high METEOR and CLIPScore values. Specifically, $\mathcal{S}_{\text{synthetic}}$ achieves a near-ideal ratio of 1.3, low LIC of 1.2, and a METEOR score of 29.3 for BLIP-2, with similar trends observed for ClipCap and Transformer.

While LIBRA effectively reduces LIC, it shows an increase in the ratio metric, indicating a trade-off between debiasing effectiveness and caption quality. Over-sampling and sub-sampling methods result in varying degrees of performance. Sub-sampling showed improved bias metrics compared to over-sampling but results in worse METEOR scores, especially for the Transformer model.

As in the multi-label classification task, we observe that although $\mathcal{S}_{\text{augment}}$ significantly reduces bias compared to using the original dataset, there is a significant gap between it and $\mathcal{S}_{\text{synthetic}}$ in terms of bias mitigation.

\begin{table*}[t]
\centering
\scriptsize
\centering
\begin{tabularx}{0.98\textwidth}{l r r r r r r r r r r r r r r r}
\toprule
& \multicolumn{3}{c}{ResNet-50} &&\multicolumn{3}{c}{Swin-T} &&\multicolumn{3}{c}{ClipCap} &&\multicolumn{3}{c}{BLIP-2}\\ 
\cline{2-4}
\cline{6-8}
\cline{10-12}
\cline{14-16}
 & \multirow{1.3}{*}{Ratio$_{\text{orig}}$} & \multirow{1.3}{*}{Ratio$_{\text{inp}}$} & \multirow{1.3}{*}{$\Delta$} && \multirow{1.3}{*}{Ratio$_{\text{orig}}$} & \multirow{1.3}{*}{Ratio$_{\text{inp}}$} & \multirow{1.3}{*}{$\Delta$} && \multirow{1.3}{*}{Ratio$_{\text{orig}}$} & \multirow{1.3}{*}{Ratio$_{\text{inp}}$} & \multirow{1.3}{*}{$\Delta$} && \multirow{1.3}{*}{Ratio$_{\text{orig}}$} & \multirow{1.3}{*}{Ratio$_{\text{inp}}$} & \multirow{1.3}{*}{$\Delta$}\\
\midrule
Original  & 3.5 & 3.0 & 14.3 && 3.1 & 2.6 & 16.1 && 2.3 & 2.5  & 8.7 && 2.3 & 2.4 & 4.4 \\
\midrule
$\mathcal{S}_{\text{augment}}$ & 3.7 & 1.5 & 59.5 && 3.2 & 0.6 & 81.3 && 2.5  & 0.8 & 68.0 && 2.3 & 1.8 & 21.7 \\
$\mathcal{S}_{\text{synthetic}}$ & 1.9 & 1.8  & 5.3 && 2.1 & 2.0 & 4.8 && 1.7  & 1.6 & 5.9 && 1.8 & 1.7 & 5.6  \\
\bottomrule
\end{tabularx}
\vspace{-5pt}
\caption{Comparison of the original ($\text{Ratio}_{\text{orig}}$) and inpainted ($\text{Ratio}_{\text{inp}}$) versions of the COCO test set. The relative difference is denoted by $\Delta = 100 \cdot \lvert \frac{\text{Ratio}_{\text{orig}} - \text{Ratio}_{\text{inp}}}{\text{Ratio}_{\text{orig}}} \rvert \%$. A larger $\Delta$ signifies a greater change.}
\label{tab:whole-body}
\end{table*}

\subsection{Analysis of Synthetic Artifacts}
\label{sec:artifacts}
Recent studies show that text-to-image models introduce synthetic artifacts in images, which models may exploit~\cite{qraitem2023fake,corvi2023detection,wang2023dire}. Our observations in \Cref{sec:mlc,sec:ic} suggest that bias persists with $\mathcal{S}_{\text{augment}}$, which augments the dataset with counterfactual images to balance group distributions. We hypothesize that $\mathcal{S}_{\text{augment}}$ may lead to shortcut learning due to spurious correlations between minoritized groups and inpainted artifacts. In contrast, $\mathcal{S}_{\text{synthetic}}$ distributes artifacts equally across all groups, avoiding this issue. 

To test this, we create a test set by inpainting random body parts using COCO-WholeBody annotations~\cite{jin2020whole}. Given an image, its caption, and body part annotations (e.g., left hand, right hand, head), we randomly select a body part, create a mask using the Segment Anything Model~\cite{kirillov2023segment}, and perform inpainting with the caption as a prompt. We evaluate the consistency of ratios between the original and synthetic test sets; a gap indicates the exploitation of synthetic artifacts for gender prediction.

\Cref{tab:whole-body} presents scores for multi-label classification (ResNet-50, Swin-T) and image captioning (ClipCap, BLIP-2). The table includes the ratio of gender predictions (\#\texttt{man}-to-\#\texttt{woman}) for the original test set ($\text{Ratio}_{\text{orig}}$) and the inpainted test set ($\text{Ratio}_{\text{inp}}$), along with the relative difference ($\Delta$) between these ratios. Results show a significant shift in gender predictions with $\mathcal{S}_{\text{augment}}$-trained models. Despite identical gender ratios in the original and inpainted test sets (both set at 2.3), models trained with $\mathcal{S}_{\text{augment}}$ predict \texttt{woman} much more frequently for the inpainted test set, indicated by the large relative differences. In contrast, models trained solely on synthetic data ($\mathcal{S}_{\text{synthetic}}$) show minimal relative differences, indicating consistent gender predictions across original and inpainted test sets.

\begin{figure}[t]
   \centering
   \includegraphics[clip, width=0.95\columnwidth]{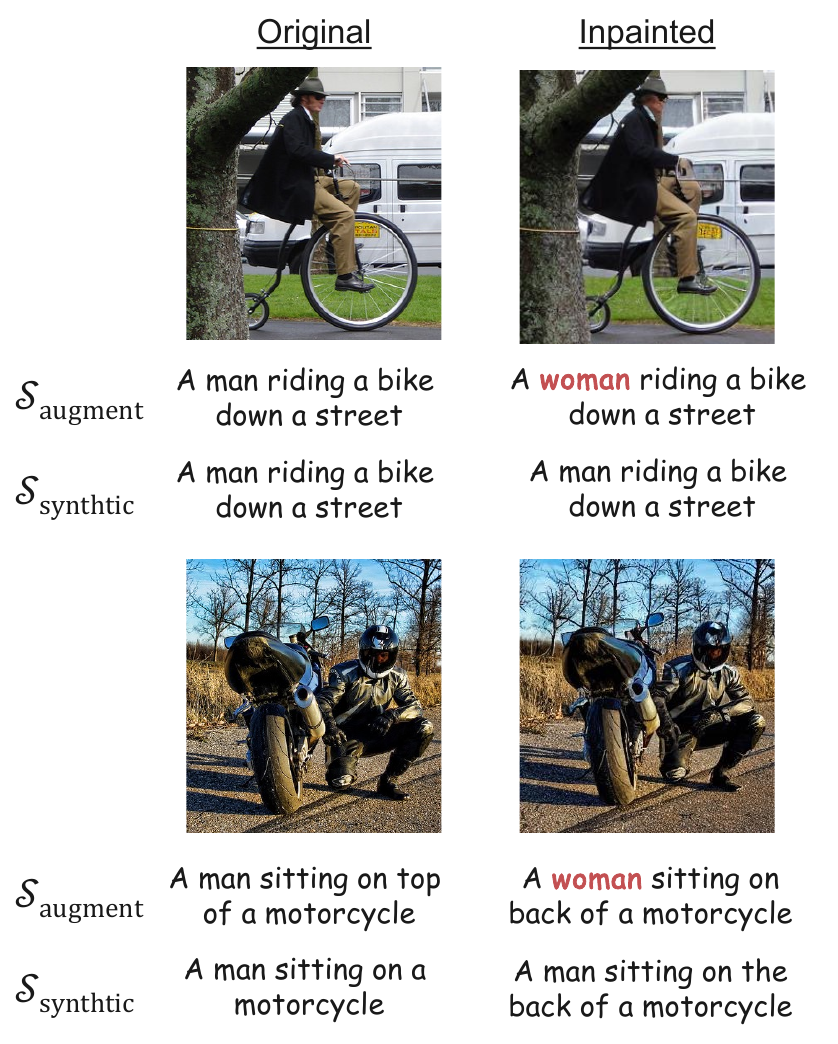}
   \vspace{-5pt}
   \caption{Predicted captions for the original (left) and inpainted (right) test images.}
   \label{fig:inp-test}
   \vspace{-3pt}
 \end{figure}

\Cref{fig:inp-test} shows examples of synthetic images and predictions by ClipCap (trained on $\mathcal{S}_{\text{augment}}$ or $\mathcal{S}_{\text{synthetic}}$). The examples demonstrate inconsistent gender predictions with $\mathcal{S}_{\text{augment}}$; specifically, the model tends to predict \texttt{woman} for the inpainted test images, evidencing exploitation of synthetic artifacts.

 \begin{table}[t]
\scriptsize
\centering
\begin{tabularx}{1\columnwidth}{X r r r r r r r}
\toprule
 & Object & Color  & Skin  & Gender & CS\\
\midrule
$s_{\text{prompt}} + s_{\text{object}} + s_{\text{color}}$       & \textbf{0.57} & 0.46 & \underline{0.29} & 0.95 & \textbf{75.3} \\
$s_{\text{prompt}} + s_{\text{object}}$ & 0.49 & 0.50 & 0.20 & \textbf{0.99} & 74.8 \\
$s_{\text{prompt}} + s_{\text{color}}$ & 0.45 & \textbf{0.56} & 0.21 & 0.94 & \underline{75.2} \\
$s_{\text{object}} + s_{\text{color}}$ & \underline{0.53} & \underline{0.52} & 0.20 & 0.96 & 74.8 \\
$s_{\text{prompt}}$       & 0.32 & 0.46 & 0.26 & \underline{0.97} & 75.1 \\
$s_{\text{object}}$       & 0.36 & 0.43 & 0.25 & 0.95 & 74.5 \\
$s_{\text{color}}$       & 0.52 & 0.50 & \textbf{0.30} & 0.95 & 74.6 \\
No filter & 0.09 & 0.07 & 0.18 & 0.94 & 74.6 \\
\bottomrule
\end{tabularx}
\vspace{-2 pt}
\caption{Human evaluation and captioning quality (CLIPScore, CS in short) for each filter combination. Higher values indicate better alignment with original images. \textbf{Bold} and \underline{underline} represent the best and second-best score for each metric.}
\label{tab:filtering}
\vspace{-2pt}
\end{table}

\begin{figure*}[t]
   \centering
   \includegraphics[clip, width=0.95\textwidth]{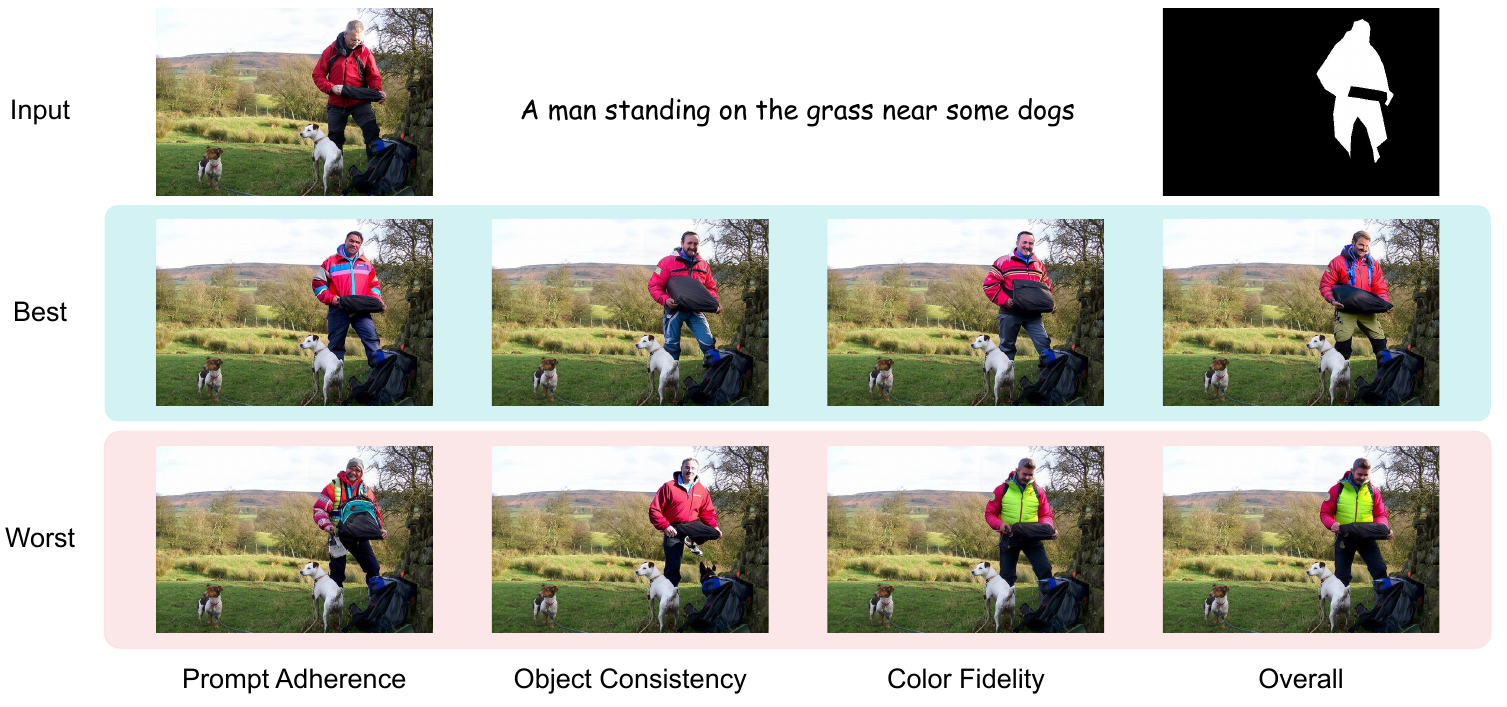}
   \caption{\hlnewblue{Best}/\hlnewred{worst} inpainted images for each filter in \Cref{sec:method-2} and their combination (overall).}
   \label{fig:filter-quali}
 \end{figure*}

\subsection{Human Filter Evaluation}
\label{sec:human}
We conduct human evaluations on Amazon Mechanical Turk~\cite{turk2012amazon} to evaluate the effectiveness of our filters, aiming to determine if our filters prevent additional biases from inpainting models and ensure high-quality images.
For 300 randomly selected original images, we analyze inpainted images chosen by each filter combination. Evaluations focus on the similarity of 1) held/nearby objects, 2) object color, and 3) skin tone compared to the original images. Workers assess differences between original and synthetic images for objects and their color, and selected skin tone classes using the Monk Skin Tone Scale~\cite{schumann2023consensus,monk2023monk}. Additionally, workers verify accurate gender depiction through a sentence gap-filling exercise (e.g., ``\texttt{A \_\_\_\_ with a dog.}''), where they must choose a protected group term to complete the sentence. More details are in \Cref{sec:app-human}.

For the evaluation of the similarity of objects and their colors, scores are computed as the proportion of times the inpainted images are rated as similar. Regarding the skin tone and gender evaluations, the scores are calculated as the proportion of matching responses from workers between the original and inpainted images. All the scores range from 0 to 1.

\Cref{tab:filtering} summarizes the human evaluation and captioning performance of ClipCap trained on $\mathcal{S}_{\text{synthetic}}$ (CS), with images selected by each filter. Notably, using all filters consistently received higher ratings across most criteria. In contrast, randomly selecting images without any filtering often leads to synthetic images differing significantly from the originals. This indicates that our filters are effective in mitigating additional biases introduced by the inpainting model. Furthermore, CLIPScore shows that using all filters improves captioning performance, highlighting its effectiveness in selecting higher-quality images.

\subsection{Inherited Biases}
\label{sec:inherited}
To further discuss the potential biases introduced by the models used in our method, we conduct several assessments.
First, for the object detector, we run Detic \cite{zhou2022detecting} on both real and synthetic images, achieving similar mAP scores of 32.0 for real images and 32.3 for synthetic images, indicating consistent performance.
Second, addressing biases in CLIP, we acknowledge the potential biases inherent in the model. However, our use of object- and color-based filters helps mitigate these biases. Additionally, image classification and captioning results verify that our method effectively reduces gender and skin tone biases.
Lastly, for the inpainting model, our filters effectively remove synthetic images that deviate from the prompt, alter color statistics, or introduce undescribed objects, as shown in \Cref{tab:filtering}. These assessments confirm that our method successfully mitigates biases without compromising performance.

\subsection{Qualitative Results}
\label{sec:qualitative}
We present qualitative examples of bias mitigation by applying our method ($\mathcal{S}_{\text{synthetic}}$) in \Cref{fig:fig1-quali}. The results show that training models on $\mathcal{S}_{\text{synthetic}}$ produces less biased outputs. For instance, in the classification task, the baseline ResNet-50 model and the over-sampling model incorrectly predict \texttt{tie}, due to its frequent co-occurrence with \texttt{man} in the training set. In contrast, $\mathcal{S}_{\text{synthetic}}$ results in a gender bias-free prediction. Image captioning results further validate our approach. The baseline ClipCap model and LIBRA model generate the man-stereotypical word \texttt{skateboard}, whereas our method correctly predicts the object \texttt{frisbee}.

In \Cref{fig:filter-quali}, we also present the best and worst inpainted images for each filter (prompt adherence, object consistency, and color fidelity), as well as their combination (overall). The results demonstrate each filter's effectiveness, and combining them selects a high-quality image that closely resembles the original. For instance, the image judged worst by the object consistency filter lacks the object the \texttt{man} is holding, while the color fidelity filter's worst image shows significant color changes in the \texttt{man}'s clothing. Combining these filters helps select an inpainted image that minimizes additional bias and closely matches the original.

\section{Conclusion}
\label{sec:conclusion}
We present a dataset-level bias mitigation pipeline that effectively reduces gender and skin tone biases by ensuring group-independent attribute distribution using synthetic-only images. Our findings indicate that mixing real and synthetic images introduces spurious correlations, underscoring the need for caution when augmenting datasets with synthetic data. Our work highlights the potential of synthetic data in bias mitigation and suggests further exploration into optimizing synthetic data generation and integration techniques for increased bias reduction.

\section*{Limitations}
\label{sec:limitations}

\paragraph{Binarized Group Classes and Intersectional Bias Analysis.}
While acknowledging that gender and skin tone exist on a spectrum, our data limitations necessitated a focus on binarized groups (i.e., \texttt{man}, \texttt{woman}), similar to prior work~\cite{zhao2017mals,wang2019balanced,zhao2023men,zhao2021captionbias}. Our analysis centered on gender and skin tone separately. However, our method can be extended to handle intersectional attributes (e.g., gender and skin tone) by inpainting with combinations of attributes (e.g., \{\texttt{woman}, \texttt{darker-skinned}\}, \{\texttt{woman}, \texttt{lighter-skinned}\}, \{\texttt{man}, \texttt{darker-skinned}\}, \{\texttt{man}, \texttt{lighter-skinned}\}). We leave this extension for future work to ensure a more comprehensive and inclusive analysis of biases.

\paragraph{Risks of Using Pre-trained Models.}
As discussed in \Cref{sec:inherited}, the pre-trained models employed in our framework (e.g., inpainting model, object detector) may introduce inherent biases. While our analysis in \Cref{sec:inherited} confirmed that these models do not adversely affect our method based on our evaluations, it is possible that some biases were not detected. Future research should focus on incorporating additional filters to further mitigate risks associated with pre-trained models.

\paragraph{Residual Bias.}
Our experimental results demonstrated that our method significantly mitigates societal bias compared to existing methods. However, bias is not completely eliminated (e.g., leakage is not zero). Future work could explore further debiasing by optimizing the weight of each filter (currently, all filters are equally weighted), introducing additional filters, and combining our method with existing bias mitigation techniques (e.g., focal loss).

\paragraph{Extending to Additional Protected Groups.}
Due to a lack of annotations for other protected attributes, our focus in this paper is on gender and skin tone biases. Nevertheless, our pipeline is applicable to various protected attributes, such as age (e.g., ``\texttt{A \textbf{woman} with a dog}'' $\rightarrow$ ``\texttt{An \textbf{elderly} woman with a dog}''). Future research should explore the application of our method to additional protected attributes.

\section*{Ethics Statement}
\label{sec:ethics}
Our research involves the manipulation of image data to mitigate societal bias, raising important ethical considerations. We address these concerns by creating synthetic images that completely inpaint over identifiable individuals, thereby respecting privacy and consent without altering their appearance. Our approach aims to promote fairness and equity by ensuring diverse and unbiased representation in image datasets. We acknowledge the potential biases inherent in the pre-trained models used and have implemented filters to mitigate these biases as much as possible. Future work should continue to explore ethical guidelines and safeguards to ensure the responsible use of generative models in research.

\section*{Acknowledgments}
This work was funded by Sony Research. Additionally, this work was partially supported by CREST Grant No. JPMJCR20D3, JST FOREST Grant No. JPMJFR216O, and JSPS KAKENHI No. JP23H00497.

\bibliography{references}

\begin{thebibliography}{81}
\providecommand{\natexlab}[1]{#1}

\bibitem[{Agarwal et~al.(2022)Agarwal, Muku, Anand, and Arora}]{agarwal2022does}
Sharat Agarwal, Sumanyu Muku, Saket Anand, and Chetan Arora. 2022.
\newblock Does data repair lead to fair models? curating contextually fair data to reduce model bias.
\newblock In \emph{WACV}.

\bibitem[{Alvi et~al.(2018)Alvi, Zisserman, and Nell{\aa}ker}]{alvi2018turning}
Mohsan Alvi, Andrew Zisserman, and Christoffer Nell{\aa}ker. 2018.
\newblock Turning a blind eye: Explicit removal of biases and variation from deep neural network embeddings.
\newblock In \emph{ECCV Workshops}.

\bibitem[{Andrews et~al.(2023)Andrews, Zhao, Thong, Modas, Papakyriakopoulos, and Xiang}]{andrews2023ethical}
Jerone Andrews, Dora Zhao, William Thong, Apostolos Modas, Orestis Papakyriakopoulos, and Alice Xiang. 2023.
\newblock Ethical considerations for responsible data curation.
\newblock In \emph{Thirty-seventh Conference on Neural Information Processing Systems Datasets and Benchmarks Track}.

\bibitem[{Bansal et~al.(2022)Bansal, Yin, Monajatipoor, and Chang}]{bansal2022well}
Hritik Bansal, Da~Yin, Masoud Monajatipoor, and Kai-Wei Chang. 2022.
\newblock How well can text-to-image generative models understand ethical natural language interventions?
\newblock In \emph{EMNLP}.

\bibitem[{Bianchi et~al.(2023)Bianchi, Kalluri, Durmus, Ladhak, Cheng, Nozza, Hashimoto, Jurafsky, Zou, and Caliskan}]{bianchi2023easily}
Federico Bianchi, Pratyusha Kalluri, Esin Durmus, Faisal Ladhak, Myra Cheng, Debora Nozza, Tatsunori Hashimoto, Dan Jurafsky, James Zou, and Aylin Caliskan. 2023.
\newblock Easily accessible text-to-image generation amplifies demographic stereotypes at large scale.
\newblock In \emph{FAccT}.

\bibitem[{Birhane et~al.(2024)Birhane, Han, Boddeti, Luccioni et~al.}]{birhane2024into}
Abeba Birhane, Sanghyun Han, Vishnu Boddeti, Sasha Luccioni, et~al. 2024.
\newblock Into the laion’s den: Investigating hate in multimodal datasets.
\newblock \emph{Advances in Neural Information Processing Systems Datasets and Benchmarks Track (NeurIPS D\&B)}.

\bibitem[{Birhane and Prabhu(2021)}]{prabhu2020large}
Abeba Birhane and Vinay~Uday Prabhu. 2021.
\newblock Large image datasets: A pyrrhic win for computer vision?
\newblock In \emph{WACV}.

\bibitem[{Birhane et~al.(2021)Birhane, Prabhu, and Kahembwe}]{birhane2021multimodal}
Abeba Birhane, Vinay~Uday Prabhu, and Emmanuel Kahembwe. 2021.
\newblock Multimodal datasets: Misogyny, pornography, and malignant stereotypes.
\newblock \emph{arXiv preprint arXiv:2110.01963}.

\bibitem[{Brinkmann et~al.(2023)Brinkmann, Swoboda, and Bartelt}]{brinkmann2023multidimensional}
Jannik Brinkmann, Paul Swoboda, and Christian Bartelt. 2023.
\newblock A multidimensional analysis of social biases in vision transformers.
\newblock In \emph{ICCV}.

\bibitem[{Brooks et~al.(2023)Brooks, Holynski, and Efros}]{brooks2023instructpix2pix}
Tim Brooks, Aleksander Holynski, and Alexei~A Efros. 2023.
\newblock Instructpix2pix: Learning to follow image editing instructions.
\newblock In \emph{CVPR}.

\bibitem[{Byrd and Lipton(2019)}]{byrd2019effect}
Jonathon Byrd and Zachary Lipton. 2019.
\newblock What is the effect of importance weighting in deep learning?
\newblock In \emph{ICML}.

\bibitem[{Cho et~al.(2023)Cho, Zala, and Bansal}]{cho2023dall}
Jaemin Cho, Abhay Zala, and Mohit Bansal. 2023.
\newblock Dall-eval: Probing the reasoning skills and social biases of text-to-image generation models.
\newblock In \emph{ICCV}.

\bibitem[{Corvi et~al.(2023)Corvi, Cozzolino, Zingarini, Poggi, Nagano, and Verdoliva}]{corvi2023detection}
Riccardo Corvi, Davide Cozzolino, Giada Zingarini, Giovanni Poggi, Koki Nagano, and Luisa Verdoliva. 2023.
\newblock On the detection of synthetic images generated by diffusion models.
\newblock In \emph{ICASSP}.

\bibitem[{Couairon et~al.(2023)Couairon, Verbeek, Schwenk, and Cord}]{couairon2022diffedit}
Guillaume Couairon, Jakob Verbeek, Holger Schwenk, and Matthieu Cord. 2023.
\newblock Diffedit: Diffusion-based semantic image editing with mask guidance.
\newblock In \emph{ICLR}.

\bibitem[{Cui et~al.(2019)Cui, Jia, Lin, Song, and Belongie}]{cui2019class}
Yin Cui, Menglin Jia, Tsung-Yi Lin, Yang Song, and Serge Belongie. 2019.
\newblock Class-balanced loss based on effective number of samples.
\newblock In \emph{CVPR}.

\bibitem[{de~Vries et~al.(2019)de~Vries, Misra, Wang, and van~der Maaten}]{de2019does}
Terrance de~Vries, Ishan Misra, Changhan Wang, and Laurens van~der Maaten. 2019.
\newblock Does object recognition work for everyone?
\newblock In \emph{CVPR Workshops}.

\bibitem[{DeVries et~al.(2019)DeVries, Misra, Wang, and van~der Maaten}]{devries2019everyone}
Terrance DeVries, Ishan Misra, Changhan Wang, and Laurens van~der Maaten. 2019.
\newblock Does object recognition work for everyone?
\newblock In \emph{CVPR Workshop on Fairness, Accountability Transparency, and Ethics in Computer Vision}.

\bibitem[{Dosovitskiy et~al.(2021)Dosovitskiy, Beyer, Kolesnikov, Weissenborn, Zhai, Unterthiner, Dehghani, Minderer, Heigold, Gelly et~al.}]{dosovitskiy2020image}
Alexey Dosovitskiy, Lucas Beyer, Alexander Kolesnikov, Dirk Weissenborn, Xiaohua Zhai, Thomas Unterthiner, Mostafa Dehghani, Matthias Minderer, Georg Heigold, Sylvain Gelly, et~al. 2021.
\newblock An image is worth 16x16 words: Transformers for image recognition at scale.
\newblock In \emph{ICLR}.

\bibitem[{Friedrich et~al.(2023)Friedrich, Schramowski, Brack, Struppek, Hintersdorf, Luccioni, and Kersting}]{friedrich2023fair}
Felix Friedrich, Patrick Schramowski, Manuel Brack, Lukas Struppek, Dominik Hintersdorf, Sasha Luccioni, and Kristian Kersting. 2023.
\newblock Fair diffusion: Instructing text-to-image generation models on fairness.
\newblock \emph{arXiv preprint arXiv:2302.10893}.

\bibitem[{Garcia et~al.(2023)Garcia, Hirota, Wu, and Nakashima}]{garcia2023uncurated}
Noa Garcia, Yusuke Hirota, Yankun Wu, and Yuta Nakashima. 2023.
\newblock Uncurated image-text datasets: Shedding light on demographic bias.
\newblock In \emph{Proceedings of the IEEE/CVF Conference on Computer Vision and Pattern Recognition}, pages 6957--6966.

\bibitem[{He et~al.(2016)He, Zhang, Ren, and Sun}]{resnet}
Kaiming He, Xiangyu Zhang, Shaoqing Ren, and Jian Sun. 2016.
\newblock Identity mappings in deep residual networks.
\newblock In \emph{ECCV}.

\bibitem[{Hendricks et~al.(2018)Hendricks, Burns, Saenko, Darrell, and Rohrbach}]{hendricks2018women}
Lisa~Anne Hendricks, Kaylee Burns, Kate Saenko, Trevor Darrell, and Anna Rohrbach. 2018.
\newblock Women also snowboard: Overcoming bias in captioning models.
\newblock In \emph{ECCV}.

\bibitem[{Hessel et~al.(2021)Hessel, Holtzman, Forbes, Bras, and Choi}]{hessel2021clipscore}
Jack Hessel, Ari Holtzman, Maxwell Forbes, Ronan~Le Bras, and Yejin Choi. 2021.
\newblock Clipscore: A reference-free evaluation metric for image captioning.
\newblock In \emph{EMNLP}.

\bibitem[{Hirota et~al.(2022)Hirota, Nakashima, and Garcia}]{hirota2022quantifying}
Yusuke Hirota, Yuta Nakashima, and Noa Garcia. 2022.
\newblock Quantifying societal bias amplification in image captioning.
\newblock In \emph{CVPR}.

\bibitem[{Hirota et~al.(2023)Hirota, Nakashima, and Garcia}]{hirota2023model}
Yusuke Hirota, Yuta Nakashima, and Noa Garcia. 2023.
\newblock Model-agnostic gender debiased image captioning.
\newblock In \emph{CVPR}.

\bibitem[{Jin et~al.(2020)Jin, Xu, Xu, Wang, Liu, Qian, Ouyang, and Luo}]{jin2020whole}
Sheng Jin, Lumin Xu, Jin Xu, Can Wang, Wentao Liu, Chen Qian, Wanli Ouyang, and Ping Luo. 2020.
\newblock Whole-body human pose estimation in the wild.
\newblock In \emph{ECCV}.

\bibitem[{Joo and K{\"a}rkk{\"a}inen(2020)}]{joo2020gender}
Jungseock Joo and Kimmo K{\"a}rkk{\"a}inen. 2020.
\newblock Gender slopes: Counterfactual fairness for computer vision models by attribute manipulation.
\newblock In \emph{International Workshop on Fairness, Accountability, Transparency and Ethics in Multimedia (FATE/MM)}.

\bibitem[{Kaushik et~al.(2019)Kaushik, Hovy, and Lipton}]{kaushik2019learning}
Divyansh Kaushik, Eduard Hovy, and Zachary Lipton. 2019.
\newblock Learning the difference that makes a difference with counterfactually-augmented data.
\newblock In \emph{ICLR}.

\bibitem[{Kingma and Ba(2015)}]{kingma2015adam}
Diederik~P. Kingma and Jimmy Ba. 2015.
\newblock Adam: A method for stochastic optimization.
\newblock In \emph{ICLR}.

\bibitem[{Kirillov et~al.(2023)Kirillov, Mintun, Ravi, Mao, Rolland, Gustafson, Xiao, Whitehead, Berg, Lo et~al.}]{kirillov2023segment}
Alexander Kirillov, Eric Mintun, Nikhila Ravi, Hanzi Mao, Chloe Rolland, Laura Gustafson, Tete Xiao, Spencer Whitehead, Alexander~C Berg, Wan-Yen Lo, et~al. 2023.
\newblock Segment anything.
\newblock \emph{arXiv preprint arXiv:2304.02643}.

\bibitem[{Krasin et~al.(2017)Krasin, Duerig, Alldrin, Ferrari, Abu-El-Haija, Kuznetsova, Rom, Uijlings, Popov, Veit et~al.}]{krasin2017openimages}
Ivan Krasin, Tom Duerig, Neil Alldrin, Vittorio Ferrari, Sami Abu-El-Haija, Alina Kuznetsova, Hassan Rom, Jasper Uijlings, Stefan Popov, Andreas Veit, et~al. 2017.
\newblock Openimages: A public dataset for large-scale multi-label and multi-class image classification.
\newblock \emph{Dataset available from https://github. com/openimages}.

\bibitem[{Li et~al.(2023)Li, Li, Savarese, and Hoi}]{li2023blip}
Junnan Li, Dongxu Li, Silvio Savarese, and Steven Hoi. 2023.
\newblock Blip-2: Bootstrapping language-image pre-training with frozen image encoders and large language models.
\newblock \emph{arXiv preprint arXiv:2301.12597}.

\bibitem[{Lin et~al.(2017)Lin, Goyal, Girshick, He, and Doll{\'a}r}]{lin2017focal}
Tsung-Yi Lin, Priya Goyal, Ross Girshick, Kaiming He, and Piotr Doll{\'a}r. 2017.
\newblock Focal loss for dense object detection.
\newblock In \emph{ICCV}.

\bibitem[{Lin et~al.(2014)Lin, Maire, Belongie, Hays, Perona, Ramanan, Doll{\'a}r, and Zitnick}]{lin2014microsoft}
Tsung-Yi Lin, Michael Maire, Serge Belongie, James Hays, Pietro Perona, Deva Ramanan, Piotr Doll{\'a}r, and C~Lawrence Zitnick. 2014.
\newblock Microsoft {COCO}: Common objects in context.
\newblock In \emph{ECCV}.

\bibitem[{Liu et~al.(2021)Liu, Lin, Cao, Hu, Wei, Zhang, Lin, and Guo}]{liu2021swin}
Ze~Liu, Yutong Lin, Yue Cao, Han Hu, Yixuan Wei, Zheng Zhang, Stephen Lin, and Baining Guo. 2021.
\newblock Swin transformer: Hierarchical vision transformer using shifted windows.
\newblock In \emph{ICCV}.

\bibitem[{Liu et~al.(2022)Liu, Mao, Wu, Feichtenhofer, Darrell, and Xie}]{liu2022convnet}
Zhuang Liu, Hanzi Mao, Chao-Yuan Wu, Christoph Feichtenhofer, Trevor Darrell, and Saining Xie. 2022.
\newblock A convnet for the 2020s.
\newblock In \emph{CVPR}.

\bibitem[{Loshchilov and Hutter(2019)}]{loshchilov2017decoupled}
Ilya Loshchilov and Frank Hutter. 2019.
\newblock Decoupled weight decay regularization.
\newblock In \emph{ICLR}.

\bibitem[{Luccioni et~al.(2023)Luccioni, Akiki, Mitchell, and Jernite}]{luccioni2023stable}
Alexandra~Sasha Luccioni, Christopher Akiki, Margaret Mitchell, and Yacine Jernite. 2023.
\newblock Stable bias: Analyzing societal representations in diffusion models.
\newblock In \emph{NeurIPS}.

\bibitem[{Mandal et~al.(2023)Mandal, Leavy, and Little}]{mandal2023multimodal}
Abhishek Mandal, Susan Leavy, and Suzanne Little. 2023.
\newblock Multimodal composite association score: Measuring gender bias in generative multimodal models.
\newblock \emph{arXiv preprint arXiv:2304.13855}.

\bibitem[{Meister et~al.(2023)Meister, Zhao, Wang, Ramaswamy, Fong, and Russakovsky}]{meister2023gender}
Nicole Meister, Dora Zhao, Angelina Wang, Vikram~V Ramaswamy, Ruth Fong, and Olga Russakovsky. 2023.
\newblock Gender artifacts in visual datasets.
\newblock In \emph{ICCV}.

\bibitem[{Misra et~al.(2016)Misra, Lawrence~Zitnick, Mitchell, and Girshick}]{misra2016seeing}
Ishan Misra, C~Lawrence~Zitnick, Margaret Mitchell, and Ross Girshick. 2016.
\newblock Seeing through the human reporting bias: Visual classifiers from noisy human-centric labels.
\newblock In \emph{CVPR}.

\bibitem[{Mokady et~al.(2021)Mokady, Hertz, and Bermano}]{mokady2021clipcap}
Ron Mokady, Amir Hertz, and Amit~H Bermano. 2021.
\newblock Clipcap: Clip prefix for image captioning.
\newblock \emph{arXiv preprint arXiv:2111.09734}.

\bibitem[{Monk(2023)}]{monk2023monk}
Ellis Monk. 2023.
\newblock The monk skin tone scale.

\bibitem[{Naik and Nushi(2023)}]{naik2023social}
Ranjita Naik and Besmira Nushi. 2023.
\newblock Social biases through the text-to-image generation lens.
\newblock In \emph{AIES}.

\bibitem[{Oh et~al.(2016)Oh, Benenson, Fritz, and Schiele}]{oh2016faceless}
Seong~Joon Oh, Rodrigo Benenson, Mario Fritz, and Bernt Schiele. 2016.
\newblock Faceless person recognition: Privacy implications in social media.
\newblock In \emph{European Conference on Computer Vision (ECCV)}, pages 19--35. Springer.

\bibitem[{Orekondy et~al.(2018)Orekondy, Fritz, and Schiele}]{orekondy2018connecting}
Tribhuvanesh Orekondy, Mario Fritz, and Bernt Schiele. 2018.
\newblock Connecting pixels to privacy and utility: Automatic redaction of private information in images.
\newblock In \emph{IEEE/CVF Conference on Computer Vision and Pattern Recognition (CVPR)}, pages 8466--8475.

\bibitem[{Qraitem et~al.(2023)Qraitem, Saenko, and Plummer}]{qraitem2023fake}
Maan Qraitem, Kate Saenko, and Bryan~A Plummer. 2023.
\newblock From fake to real (ffr): A two-stage training pipeline for mitigating spurious correlations with synthetic data.
\newblock \emph{arXiv preprint arXiv:2308.04553}.

\bibitem[{Radford et~al.(2021)Radford, Kim, Hallacy, Ramesh, Goh, Agarwal, Sastry, Askell, Mishkin, Clark et~al.}]{radford2021learning}
Alec Radford, Jong~Wook Kim, Chris Hallacy, Aditya Ramesh, Gabriel Goh, Sandhini Agarwal, Girish Sastry, Amanda Askell, Pamela Mishkin, Jack Clark, et~al. 2021.
\newblock Learning transferable visual models from natural language supervision.
\newblock In \emph{ICML}.

\bibitem[{Radford et~al.(2019)Radford, Wu, Child, Luan, Amodei, Sutskever et~al.}]{radford2019language}
Alec Radford, Jeffrey Wu, Rewon Child, David Luan, Dario Amodei, Ilya Sutskever, et~al. 2019.
\newblock Language models are unsupervised multitask learners.
\newblock \emph{OpenAI blog}.

\bibitem[{Ramaswamy et~al.(2021{\natexlab{a}})Ramaswamy, Kim, and Russakovsky}]{ramaswamy2020debiasing}
Vikram~V. Ramaswamy, Sunnie S.~Y. Kim, and Olga Russakovsky. 2021{\natexlab{a}}.
\newblock Fair attribute classification through latent space de-biasing.
\newblock In \emph{IEEE/CVF Conference on Computer Vision and Pattern Recognition (CVPR)}.

\bibitem[{Ramaswamy et~al.(2021{\natexlab{b}})Ramaswamy, Kim, and Russakovsky}]{ramaswamy2021fair}
Vikram~V Ramaswamy, Sunnie~SY Kim, and Olga Russakovsky. 2021{\natexlab{b}}.
\newblock Fair attribute classification through latent space de-biasing.
\newblock In \emph{CVPR}.

\bibitem[{Rombach et~al.(2022)Rombach, Blattmann, Lorenz, Esser, and Ommer}]{rombach2022high}
Robin Rombach, Andreas Blattmann, Dominik Lorenz, Patrick Esser, and Bj{\"o}rn Ommer. 2022.
\newblock High-resolution image synthesis with latent diffusion models.
\newblock In \emph{CVPR}.

\bibitem[{Ross et~al.(2020)Ross, Katz, and Barbu}]{ross2020measuring}
Candace Ross, Boris Katz, and Andrei Barbu. 2020.
\newblock Measuring social biases in grounded vision and language embeddings.
\newblock \emph{arXiv preprint arXiv:2002.08911}.

\bibitem[{Russakovsky et~al.(2015)Russakovsky, Deng, Su, Krause, Satheesh, Ma, Huang, Karpathy, Khosla, Bernstein, Berg, and Fei-Fei}]{imagenet}
Olga Russakovsky, Jia Deng, Hao Su, Jonathan Krause, Sanjeev Satheesh, Sean Ma, Zhiheng Huang, Andrej Karpathy, Aditya Khosla, Michael Bernstein, Alexander~C. Berg, and Li~Fei-Fei. 2015.
\newblock {ImageNet Large Scale Visual Recognition Challenge}.
\newblock \emph{IJCV}.

\bibitem[{Sagawa et~al.(2019)Sagawa, Koh, Hashimoto, and Liang}]{sagawa2019distributionally}
Shiori Sagawa, Pang~Wei Koh, Tatsunori~B Hashimoto, and Percy Liang. 2019.
\newblock Distributionally robust neural networks for group shifts: On the importance of regularization for worst-case generalization.
\newblock In \emph{ICLR}.

\bibitem[{Sattigeri et~al.(2019)Sattigeri, Hoffman, Chenthamarakshan, and Varshney}]{sattigeri2019fairness}
Prasanna Sattigeri, Samuel~C Hoffman, Vijil Chenthamarakshan, and Kush~R Varshney. 2019.
\newblock Fairness gan: Generating datasets with fairness properties using a generative adversarial network.
\newblock \emph{IBM Journal of Research and Development}.

\bibitem[{Schumann et~al.(2023)Schumann, Olanubi, Wright, Monk~Jr, Heldreth, and Ricco}]{schumann2023consensus}
Candice Schumann, Gbolahan~O Olanubi, Auriel Wright, Ellis Monk~Jr, Courtney Heldreth, and Susanna Ricco. 2023.
\newblock Consensus and subjectivity of skin tone annotation for ml fairness.
\newblock \emph{arXiv preprint arXiv:2305.09073}.

\bibitem[{Schumann et~al.(2021)Schumann, Ricco, Prabhu, Ferrari, and Pantofaru}]{schumann2021step}
Candice Schumann, Susanna Ricco, Utsav Prabhu, Vittorio Ferrari, and Caroline Pantofaru. 2021.
\newblock A step toward more inclusive people annotations for fairness.
\newblock In \emph{AIES}.

\bibitem[{Seshadri et~al.(2023)Seshadri, Singh, and Elazar}]{seshadri2023bias}
Preethi Seshadri, Sameer Singh, and Yanai Elazar. 2023.
\newblock The bias amplification paradox in text-to-image generation.
\newblock \emph{arXiv preprint arXiv:2308.00755}.

\bibitem[{Sharmanska et~al.(2020)Sharmanska, Hendricks, Darrell, and Quadrianto}]{sharmanska2020contrastive}
Viktoriia Sharmanska, Lisa~Anne Hendricks, Trevor Darrell, and Novi Quadrianto. 2020.
\newblock Contrastive examples for addressing the tyranny of the majority.
\newblock \emph{arXiv preprint arXiv:2004.06524}.

\bibitem[{Smith et~al.(2023)Smith, Farinha, Hall, Kirk, Shtedritski, and Bain}]{smith2023balancing}
Brandon Smith, Miguel Farinha, Siobhan~Mackenzie Hall, Hannah~Rose Kirk, Aleksandar Shtedritski, and Max Bain. 2023.
\newblock Balancing the picture: Debiasing vision-language datasets with synthetic contrast sets.
\newblock \emph{arXiv preprint arXiv:2305.15407}.

\bibitem[{Sobel(2020)}]{sobel2020taxonomy}
Benjamin Sobel. 2020.
\newblock A taxonomy of training data: Disentangling the mismatched rights, remedies, and rationales for restricting machine learning.
\newblock \emph{Artificial Intelligence and Intellectual Property (Reto Hilty, Jyh-An Lee, Kung-Chung Liu, eds.), Oxford University Press, Forthcoming}.

\bibitem[{Sokolova et~al.(2006)Sokolova, Japkowicz, and Szpakowicz}]{sokolova2006beyond}
Marina Sokolova, Nathalie Japkowicz, and Stan Szpakowicz. 2006.
\newblock Beyond accuracy, f-score and roc: a family of discriminant measures for performance evaluation.
\newblock In \emph{Australasian joint conference on artificial intelligence}.

\bibitem[{Struppek et~al.(2022)Struppek, Hintersdorf, and Kersting}]{struppek2022biased}
Lukas Struppek, Dominik Hintersdorf, and Kristian Kersting. 2022.
\newblock The biased artist: Exploiting cultural biases via homoglyphs in text-guided image generation models.
\newblock \emph{arXiv preprint arXiv:2209.08891}.

\bibitem[{Tang et~al.(2021)Tang, Du, Li, Liu, Zou, and Hu}]{tang2021mitigating}
Ruixiang Tang, Mengnan Du, Yuening Li, Zirui Liu, Na~Zou, and Xia Hu. 2021.
\newblock Mitigating gender bias in captioning systems.
\newblock In \emph{WWW}.

\bibitem[{Turk(2012)}]{turk2012amazon}
Amazon~Mechanical Turk. 2012.
\newblock Amazon mechanical turk.
\newblock \emph{Retrieved August}.

\bibitem[{Ungless et~al.(2023)Ungless, Ross, and Lauscher}]{ungless2023stereotypes}
Eddie~L Ungless, Bj{\"o}rn Ross, and Anne Lauscher. 2023.
\newblock Stereotypes and smut: The (mis) representation of non-cisgender identities by text-to-image models.
\newblock In \emph{ACL}.

\bibitem[{Wang et~al.(2020{\natexlab{a}})Wang, Narayanan, and Russakovsky}]{wang2020revise}
Angelina Wang, Arvind Narayanan, and Olga Russakovsky. 2020{\natexlab{a}}.
\newblock {REVISE}: A tool for measuring and mitigating bias in visual datasets.
\newblock In \emph{ECCV}.

\bibitem[{Wang and Russakovsky(2021)}]{wang2021biasamp}
Angelina Wang and Olga Russakovsky. 2021.
\newblock Directional bias amplification.
\newblock In \emph{ICML}.

\bibitem[{Wang et~al.(2023{\natexlab{a}})Wang, Liu, Di, Liu, and Wang}]{wang2023t2iat}
Jialu Wang, Xinyue~Gabby Liu, Zonglin Di, Yang Liu, and Xin~Eric Wang. 2023{\natexlab{a}}.
\newblock T2iat: Measuring valence and stereotypical biases in text-to-image generation.
\newblock In \emph{ACL}.

\bibitem[{Wang et~al.(2019)Wang, Zhao, Yatskar, Chang, and Ordonez}]{wang2019balanced}
Tianlu Wang, Jieyu Zhao, Mark Yatskar, Kai-Wei Chang, and Vicente Ordonez. 2019.
\newblock Balanced datasets are not enough: Estimating and mitigating gender bias in deep image representations.
\newblock In \emph{ICCV}.

\bibitem[{Wang et~al.(2020{\natexlab{b}})Wang, Qinami, Karakozis, Genova, Nair, Hata, and Russakovsky}]{wang2020towards}
Zeyu Wang, Klint Qinami, Ioannis~Christos Karakozis, Kyle Genova, Prem Nair, Kenji Hata, and Olga Russakovsky. 2020{\natexlab{b}}.
\newblock Towards fairness in visual recognition: Effective strategies for bias mitigation.
\newblock In \emph{CVPR}.

\bibitem[{Wang and Culotta(2021)}]{wang2021robustness}
Zhao Wang and Aron Culotta. 2021.
\newblock Robustness to spurious correlations in text classification via automatically generated counterfactuals.
\newblock In \emph{AAAI}.

\bibitem[{Wang et~al.(2023{\natexlab{b}})Wang, Bao, Zhou, Wang, Hu, Chen, and Li}]{wang2023dire}
Zhendong Wang, Jianmin Bao, Wengang Zhou, Weilun Wang, Hezhen Hu, Hong Chen, and Houqiang Li. 2023{\natexlab{b}}.
\newblock Dire for diffusion-generated image detection.
\newblock In \emph{ICCV}.

\bibitem[{Wolf et~al.(2020)Wolf, Debut, Sanh, Chaumond, Delangue, Moi, Cistac, Rault, Louf, Funtowicz et~al.}]{wolf2020transformers}
Thomas Wolf, Lysandre Debut, Victor Sanh, Julien Chaumond, Clement Delangue, Anthony Moi, Pierric Cistac, Tim Rault, R{\'e}mi Louf, Morgan Funtowicz, et~al. 2020.
\newblock Transformers: State-of-the-art natural language processing.
\newblock In \emph{EMNLP: system demonstrations}.

\bibitem[{Yew and Xiang(2022)}]{yew2022regulating}
Rui-Jie Yew and Alice Xiang. 2022.
\newblock Regulating facial processing technologies: Tensions between legal and technical considerations in the application of illinois bipa.
\newblock In \emph{ACM Conference on Fairness, Accountability, and Transparency (FAccT)}, page 1017–1027.

\bibitem[{Zhang et~al.(2023)Zhang, Jiang, Turk, and Yang}]{zhang2023auditing}
Yanzhe Zhang, Lu~Jiang, Greg Turk, and Diyi Yang. 2023.
\newblock Auditing gender presentation differences in text-to-image models.
\newblock \emph{arXiv preprint arXiv:2302.03675}.

\bibitem[{Zhao et~al.(2023)Zhao, Andrews, and Xiang}]{zhao2023men}
Dora Zhao, Jerone~TA Andrews, and Alice Xiang. 2023.
\newblock Men also do laundry: Multi-attribute bias amplification.
\newblock In \emph{ICML}.

\bibitem[{Zhao et~al.(2021)Zhao, Wang, and Russakovsky}]{zhao2021captionbias}
Dora Zhao, Angelina Wang, and Olga Russakovsky. 2021.
\newblock Understanding and evaluating racial biases in image captioning.
\newblock In \emph{ICCV}.

\bibitem[{Zhao et~al.(2017)Zhao, Wang, Yatskar, Ordonez, and Chang}]{zhao2017mals}
Jieyu Zhao, Tianlu Wang, Mark Yatskar, Vicente Ordonez, and Kai-Wei Chang. 2017.
\newblock Men also like shopping: Reducing gender bias amplification using corpus-level constraints.
\newblock In \emph{EMNLP}.

\bibitem[{Zhou et~al.(2022)Zhou, Girdhar, Joulin, Kr{\"a}henb{\"u}hl, and Misra}]{zhou2022detecting}
Xingyi Zhou, Rohit Girdhar, Armand Joulin, Philipp Kr{\"a}henb{\"u}hl, and Ishan Misra. 2022.
\newblock Detecting twenty-thousand classes using image-level supervision.
\newblock In \emph{ECCV}.

\end{thebibliography}

\appendix


    

\section{Method Details}
\label{sec:app-detail}

\subsection{Image Generation Settings}
\label{sec:image-generation-detail}
\paragraph{Selection of People for Inpainting.}
Following the previous works \cite{zhao2021captionbias,misra2016seeing}, we apply inpainting to a person with the largest bounding box. In addition, if the second largest person's box is larger than $55,000$ pixels, the region is also inpainted. For COCO, we do this by using the \texttt{person} label and corresponding bounding boxes. For OpenImages, we use person-bounding boxes presented in More Inclusive Annotations for People (MIAP) annotations \cite{schumann2021step}, then we generate person masks within the boxes using Segment Anything Model \cite{kirillov2023segment}.

\paragraph{Parameters of Image Generation.}
In \Cref{sec:method-1}, we generate $m = 30$ inpainted images for each group (e.g., $\{\texttt{woman}, \texttt{man}\}$ for binary gender). When generating the images, we use three different guidance scale parameters (7.5, 9.5, and 15.0) to generate diverse inpainted images (i.e., generating 10 images for each guidance scale). We use 6 NVIDIA A100-PCIE-40GB GPUs, resulting in a total of 72 hours to finish synthesizing images.

\subsection{Visual examples of inpainted images \& failure cases}
\begin{figure*}[t]
   \centering
   \includegraphics[clip, width=0.93\textwidth]{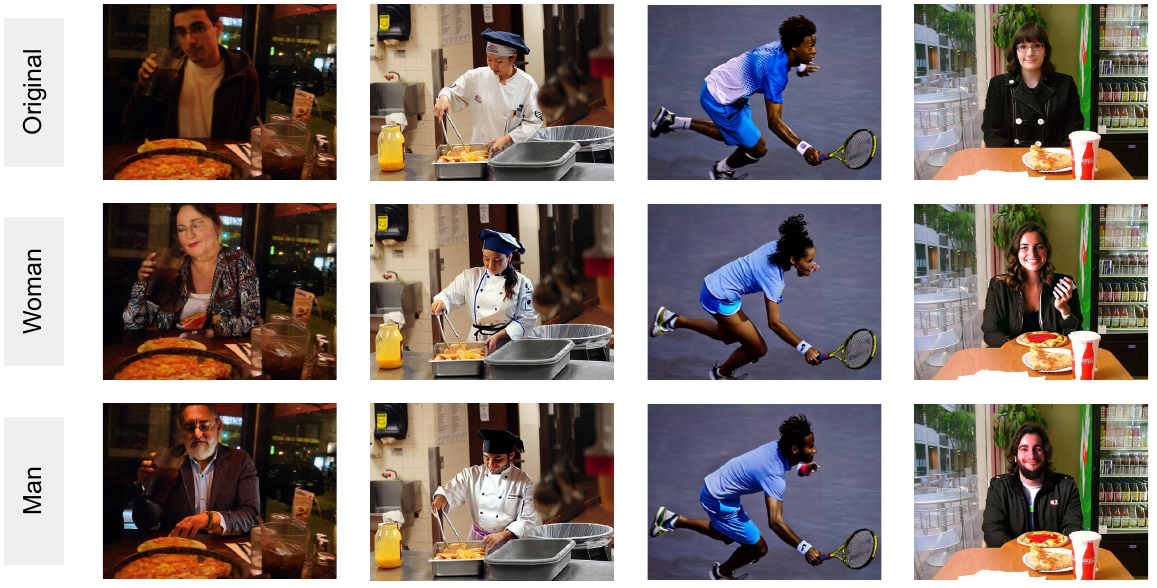}
   \caption{Examples of inpainted images for binary gender.}
   \label{fig:gender}
 \end{figure*}

\begin{figure*}[t]
   \centering
   \includegraphics[clip, width=0.94\textwidth]{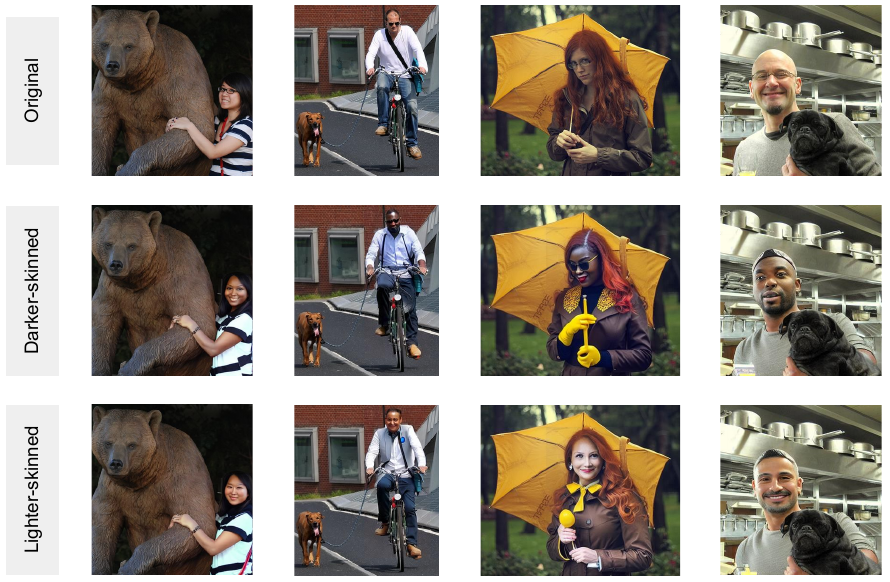}
   \caption{Examples of inpainted images for binary skin tone.}
   \label{fig:skin}
 \end{figure*}

We show the visual examples of the inpainted images after filtering in \Cref{fig:gender} (for binary gender) and \Cref{fig:skin} (for binary skin tone). 
The examples show that the inpainted images depict the target groups (e.g., \texttt{woman} and \texttt{darker-skinned}), keeping the rest fixed. In some cases, artifacts are noticeable, which enables us to identify synthetic images (e.g., the details of the faces are not clear), but they do not affect the downstream performance, as shown in the main paper. 

\begin{table*}[t]
\footnotesize
\centering
\begin{tabularx}{0.89\textwidth}{l r r r r r r r r r r r}
\toprule
& \multicolumn{3}{c}{ResNet-50} && \multicolumn{3}{c}{Swin-T} && \multicolumn{3}{c}{ConvNeXt-B} \\ 
 \cline{2-4} 
 \cline{6-8}
 \cline{10-12}
& \multirow{1.3}{*}{mAP} & \multirow{1.3}{*}{Ratio} & \multirow{1.3}{*}{Leakage} && \multirow{1.3}{*}{mAP} & \multirow{1.3}{*}{Ratio} & \multirow{1.3}{*}{Leakage}  && \multirow{1.3}{*}{mAP} & \multirow{1.3}{*}{Ratio} & \multirow{1.3}{*}{Leakage} \\
\midrule
Original  & \underline{42.3} & 5.2 & 18.9 && \underline{45.3} & 4.3 & 20.9 && \underline{46.0} & 5.0 & 22.7 \\
\midrule
Adversarial & 37.5 & — & \textbf{8.3}  && 40.8  & --- & \textbf{11.3} && 40.4 & --- & \textbf{12.3} \\
DomDisc & 40.7 & 3.7 & 20.6 && 43.6 & 4.6 & 22.1 && 42.9 & 4.1 & 21.9 \\
DomInd  & 40.3 & 3.7 & 19.1 && 42.7 & 3.5 & 20.2 && 43.4 & 2.6 & 22.0 \\
Upweight & 41.3 & 6.5 & \underline{13.1} && 44.7 & 5.8 & 17.9 && 45.3 & 7.4 & 18.0 \\
Focal  & \textbf{43.0} & 4.6 & 18.7 && \textbf{45.4} & 4.4 & 21.3 && 45.4 & 4.0 & 22.3 \\
CB  & 40.5 & 5.2 & 18.0 && 42.6 & 3.9 & 19.8 && 43.9 & 4.6 & 21.5 \\
GroupDRO  & \underline{42.3} & 4.2 & 18.9 && 45.1 & 4.2 & 20.9 && \textbf{46.1} & 3.4 & 22.5 \\
\midrule
Over-sampling & 38.5 & 3.3 & 15.0 && 41.1 & 4.0 & \underline{16.1} && 41.7 & 5.2 & 18.4 \\
Sub-sampling & 38.3 & \underline{2.2} & 18.3 && 41.2 & \underline{2.1} & 19.8 && 39.8 & 2.8 & 21.7 \\
$\mathcal{S}_{\text{augment}}$ (Ours) & 42.0 & 1.9 & 16.0 && 44.9 & 2.4 & 18.0 && 45.5 & 2.6 & 19.0 \\
$\mathcal{S}_{\text{synthetic}}$ (Ours)& 41.4 & \textbf{1.1} & 14.6 && 44.4 & \textbf{2.0} & 17.6 && 44.7 & \textbf{1.3} & \underline{17.9} \\
\bottomrule
\end{tabularx}
\caption{Classification performance and gender bias scores of ResNet-50, Swin-T, and ConvNeXt-B backbones on OpenImages. Ratio is inapplicable to Adversarial due to its gender prediction module for mitigation. \textbf{Bold} and \underline{underline} represent the best and second-best, respectively. For an unbiased model, $\text{Ratio} = \text{1}$ and $\text{Leakage} = \text{0}$.}
\label{tab:new-mlc-oi}
\end{table*}

\begin{table*}[t]
\renewcommand{\arraystretch}{1.1}
\setlength{\tabcolsep}{5pt}
\footnotesize
\centering
\begin{tabularx}{0.67\textwidth}{l r r r r r r r r}
\toprule
& \multicolumn{2}{c}{ResNet-50} &&\multicolumn{2}{c}{Swin-T} &&\multicolumn{2}{c}{ConvNeXt-B}\\ 
\cline{2-3} 
\cline{5-6}
\cline{8-9}
& mAP  & Leakage  && mAP  & Leakage  && mAP  & Leakage  \\
\midrule
Original  & $\mathbf{65.8}$ & $3.2$  && $\mathbf{72.2}$ & $7.1$  && $\mathbf{75.9}$  & $7.2$  \\
\midrule
$\mathcal{S}_{\text{synthetic}}$ (Ours) & $65.2$ & $\mathbf{2.3}$  && $71.4$  & $\mathbf{3.7}$  && $74.5$ & $\mathbf{5.9}$  \\
\bottomrule
\end{tabularx}
\caption{Classification performance and skin tone bias scores of ResNet-50, Swin-T, and ConvNeXt-B backbones on COCO. \textbf{Bold} represents the best. For an unbiased model, $\text{Ratio} = \text{1}$ and $\text{Leakage} = \text{0}$.}
\label{tab:new-mlc-skin}
\end{table*}

\section{Experimental Settings and Additional Results}
\label{sec:app-exp}

\subsection{Multi-Label Classification}
\label{sec:app-mlc}

\paragraph{Datasets.}
We use COCO \cite{lin2014microsoft} and OpenImages \cite{krasin2017openimages}. Following previous works \cite{zhao2017mals,zhao2023men}, we focus on attributes co-occurring with \texttt{woman} or \texttt{man} more than $100$ times and remove person-related classes (e.g., \texttt{person} class), resulting in $51$ and $126$ attributes for COCO and OpenImages, respectively. The list of the attributes is as follows: 

COCO: \{\texttt{sink}, \texttt{refrigerator}, \texttt{laptop}, \texttt{surfboard}, \texttt{vase}, \texttt{bottle}, \texttt{remote}, \texttt{donut}, \texttt{motorcycle}, \texttt{car}, \texttt{chair}, \texttt{suitcase}, \texttt{tv}, \texttt{knife}, \texttt{fork}, \texttt{couch}, \texttt{bus}, \texttt{toothbrush}, \texttt{bicycle}, \texttt{tie}, \texttt{clock}, \texttt{microwave}, \texttt{teddy bear}, \texttt{frisbee}, \texttt{spoon}, \texttt{dog}, \texttt{truck}, \texttt{bench}, \texttt{backpack}, \texttt{skis}, \texttt{horse}, \texttt{sandwich}, \texttt{bed}, \texttt{handbag}, \texttt{umbrella}, \texttt{pizza}, \texttt{book}, \texttt{dining table}, \texttt{traffic light}, \texttt{banana}, \texttt{potted plant}, \texttt{tennis racket}, \texttt{cat}, \texttt{sports ball}, \texttt{kite}, \texttt{cake}, \texttt{wine glass}, \texttt{bowl}, \texttt{cup}, \texttt{oven}, \texttt{cell phone}\}.

OpenImages: \{\texttt{goggles}, \texttt{building}, \texttt{cloud}, \texttt{smile}, \texttt{tree}, \texttt{sunglasses}, \texttt{light}, \texttt{t-shirt}, \texttt{glasses}, \texttt{water}, \texttt{forehead}, \texttt{wall}, \texttt{sky}, \texttt{tire}, \texttt{roof}, \texttt{road}, \texttt{wheel}, \texttt{vehicle}, \texttt{land vehicle}, \texttt{car}, \texttt{tie}, \texttt{furniture}, \texttt{microphone}, \texttt{suit}, \texttt{clothing}, \texttt{fence}, \texttt{jeans}, \texttt{trousers}, \texttt{shirt}, \texttt{footwear}, \texttt{flooring}, \texttt{outerwear}, \texttt{coat}, \texttt{ceiling}, \texttt{floor}, \texttt{jacket}, \texttt{table}, \texttt{house}, \texttt{couch}, \texttt{mammal}, \texttt{hat}, \texttt{shoe}, \texttt{sports uniform}, \texttt{baseball (sport)}, \texttt{cap}, \texttt{baseball cap}, \texttt{bag}, \texttt{drawing}, \texttt{sun hat}, \texttt{musical instrument}, \texttt{baby}, \texttt{window}, \texttt{door}, \texttt{sweater}, \texttt{lake}, \texttt{chair}, \texttt{tableware}, \texttt{bottle}, \texttt{drink}, \texttt{handwriting}, \texttt{paper}, \texttt{food}, \texttt{tent}, \texttt{concert}, \texttt{drum}, \texttt{guitar}, \texttt{glove}, \texttt{sports equipment}, \texttt{blazer}, \texttt{art}, \texttt{painting}, \texttt{dress}, \texttt{flower}, \texttt{sneakers}, \texttt{screenshot}, \texttt{watercraft}, \texttt{beach}, \texttt{animal}, \texttt{grass family}, \texttt{plant}, \texttt{soil}, \texttt{desk}, \texttt{poster}, \texttt{bus}, \texttt{computer}, \texttt{personal computer}, \texttt{watch}, \texttt{mountain}, \texttt{helmet}, \texttt{bicycle helmet}, \texttt{bicycle wheel}, \texttt{bicycle}, \texttt{curtain}, \texttt{dance}, \texttt{football}, \texttt{ball (object)}, \texttt{soccer}, \texttt{wedding dress}, \texttt{jewellery}, \texttt{bride}, \texttt{office building}, \texttt{laptop}, \texttt{toddler}, \texttt{shorts}, \texttt{hiking}, \texttt{fashion accessory}, \texttt{fedora}, \texttt{swimming}, \texttt{swimwear}, \texttt{camera}, \texttt{playground}, \texttt{weapon}, \texttt{ship}, \texttt{statue}, \texttt{boat}, \texttt{fast food}, \texttt{flag}, \texttt{soft drink}, \texttt{book}, \texttt{auto part}, \texttt{snow}, \texttt{carnivore}, \texttt{dog}, \texttt{horse}, \texttt{motorcycle}, \texttt{pole dance}\}.

\paragraph{Training.}
The models (ResNet-50~\cite{resnet}, Swin-T~\cite{liu2021swin}, and ConvNeXt-Base~\cite{liu2022convnet}) are initialized with ImageNet~\cite{imagenet} pre-training, and fine-tuned with early stopping using a validation set split from the training set ($20\%$ of the training set). The optimizer is Adam \cite{kingma2015adam}, batch size is $32$, and a learning rate is $1 \times 10^{-5}$. For binary gender, the classification layers predict both protected groups (i.e., \{\texttt{woman}, \texttt{man}\}) and object classes. For binary skin tone, the models only predict object classes as ground-truth skin tone labels are not available.

\paragraph{Results for OpenImages.}
We show the complete results of the experiments in the main paper: gender bias on OpenImages (\Cref{tab:new-mlc-oi}).
The results show that all the insights described in the main paper are consistent across the datasets.

\paragraph{Results for skin tone bias.}
Previous bias mitigation methods face a significant limitation, requiring protected group labels for all training set samples~\cite{zhao2017mals,wang2019balanced,agarwal2022does}. They typically focus on gender as a protected attribute due to its prevalence in captions \cite{misra2016seeing}, allowing for label inference through gender-related terms. In contrast, $\mathcal{S}_{\text{synthetic}}$ applies to attributes without labels, such as skin tone. We use our pipeline (excluding the color fidelity filter, as we aim to modify skin tone) on binary skin tone categories (i.e., $\mathcal{G} = \{\texttt{darker-skinned}, \texttt{lighter-skinned}\}$) using COCO. We evaluate skin tone bias using \emph{leakage} only since \emph{ratio} requires models to predict protected groups, and there are no skin tone annotations for the COCO training set.
Results are shown in \Cref{tab:new-mlc-skin}, demonstrating consistent conclusions with gender bias.


\begin{table*}[t]
\centering
\footnotesize
\begin{tabularx}{0.74\textwidth}{l r r r r r r r r r r r}
\toprule
& \multicolumn{3}{c}{ClipCap} && \multicolumn{3}{c}{BLIP-2} && \multicolumn{3}{c}{Transformer} \\ 
\cline{2-4}
\cline{6-8}
\cline{10-12}
& \multirow{1.3}{*}{M} & \multirow{1.3}{*}{CS} & \multirow{1.3}{*}{LIC} && \multirow{1.3}{*}{M} & \multirow{1.3}{*}{CS}  & \multirow{1.3}{*}{LIC} && \multirow{1.3}{*}{M} & \multirow{1.3}{*}{CS}  & \multirow{1.3}{*}{LIC} \\
\midrule
Original  & \textbf{29.4} & 75.3 & 4.6 && \textbf{27.1} & \textbf{73.9} & 2.2 && \textbf{27.0} & \textbf{71.5} & 5.3   \\
\midrule
$\mathcal{S}_{\text{synthetic}}$ (Ours)  & 29.1 & \textbf{75.4}& \textbf{3.7} && 26.8 & 73.6 & \textbf{2.0} && 26.5 & 71.0 & \textbf{4.7}   \\
\bottomrule
\end{tabularx}
\vspace{-4pt}
\caption{Captioning quality and skin tone bias scores of ClipCap, BLIP-2, and Transformer backbones on COCO. M and CS denote METEOR and CLIPScore. \textbf{Bold} represents the best. For an unbiased model, $\text{Ratio} = \text{1}$ and $\text{LIC} = \text{0}$.}
\label{tab:new-ic-skin}
\end{table*}

\subsection{Image Captioning}
\label{sec:app-ic}

\paragraph{Training.} 
We benchmark three captioning models: ClipCap \cite{mokady2021clipcap}, BLIP-2 \cite{li2023blip}, and Transformer (i.e., the Transformer-based encoder-decoder model composed of Vision Transformer \cite{dosovitskiy2020image} and GPT-2 \cite{radford2019language}). As with multi-label classification, we train the models with early stopping. Specifically, for ClipCap, we follow the official implementation regarding the training settings. For BLIP-2 and Transformer, we use the implementation in Hugging Face \cite{wolf2020transformers}. We use the AdamW optimizer \cite{loshchilov2017decoupled} with a learning rate of $2 \times 10^{-6}$/$1 \times 10^{-4}$ and batch size of $8$/$64$ for BLIP-2 and Transformer, respectively. 

\paragraph{Results for skin tone.}
We show the results of the experiments for skin tone bias mitigation in \Cref{tab:new-ic-skin}. The results show that the insights in the main paper are mostly consistent across the protected groups.

\subsection{Human Filter Evaluation}
\label{sec:app-human}
\begin{figure*}[t]
   \centering
   \includegraphics[clip, width=0.94\textwidth]{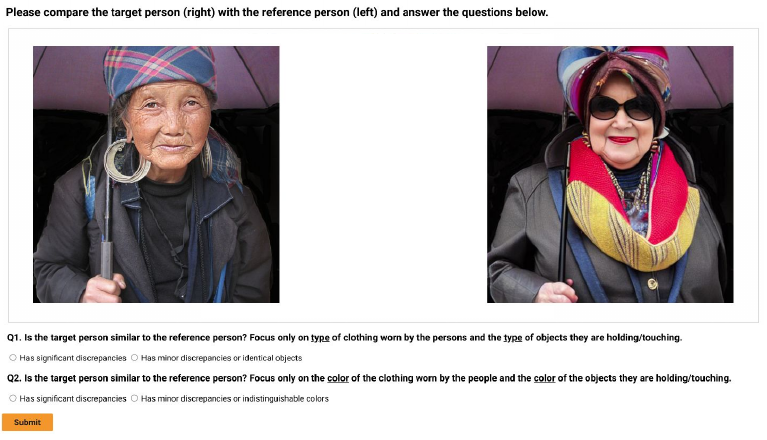}
   \caption{Evaluation of \emph{perceived} object and color similarity between original and inpainted images on AMT.}
   \label{fig:amt-img}
 \end{figure*}

\begin{figure*}[t]
   \centering
   \includegraphics[clip, width=0.94\textwidth]{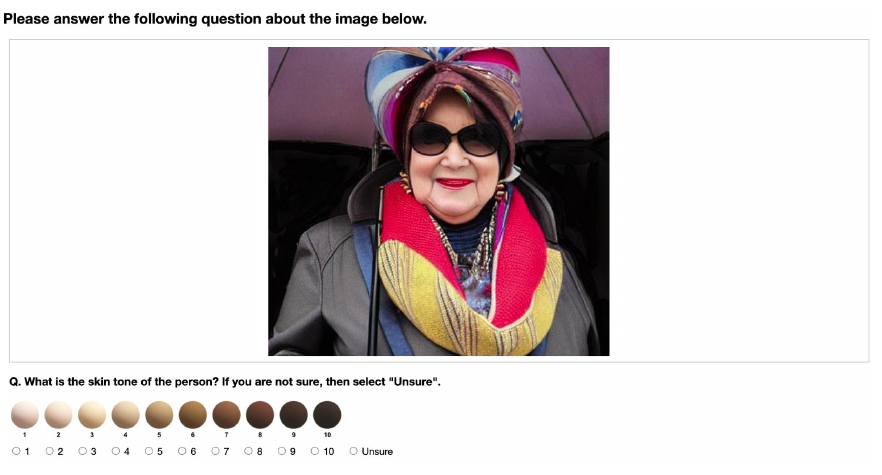}
   \caption{Evaluation of \emph{perceived} skin tone using the Monk Skin Tone Scale on AMT.}
   \label{fig:amt-skin}
 \end{figure*}

\begin{figure*}[t]
   \centering
   \includegraphics[clip, width=0.94\textwidth]{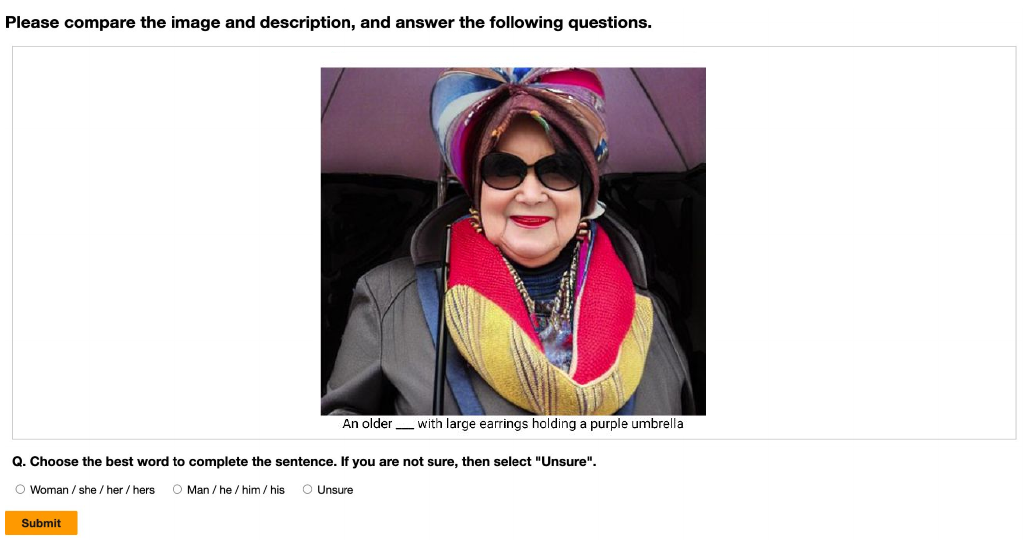}
   \caption{Evaluation of \emph{perceived} gender depiction accuracy in inpainted images on AMT.}
   \label{fig:amt-gender}
 \end{figure*}

In \Cref{fig:amt-img,fig:amt-skin,fig:amt-gender}, we present example tasks for human evaluation conducted on Amazon Mechanical Turk (AMT)~\cite{turk2012amazon}. This evaluation assesses how well each combination of filters identifies desirable inpainted images. \Cref{fig:amt-img} shows the user interface for evaluating the similarity of held/nearby objects and their colors between the original (left) and inpainted (right) images. \Cref{fig:amt-skin} asks workers to select a skin tone class using the Monk Skin Tone Scale~\cite{schumann2023consensus,monk2023monk}. We conduct this evaluation on both original and inpainted images and compute the degree of agreement between them. \Cref{fig:amt-gender} verifies if \emph{perceived} gender is accurately depicted---according to the AMT worker---in the inpainted images through gap-filling, where workers must choose a protected group term to complete the sentence. Each assignment pays \$0.07, with a total participant compensation of approximately \$2,000.

\section{Image Attribution}
All images in this paper are sourced from the COCO dataset~\cite{lin2014microsoft}, which is publicly available at \url{https://cocodataset.org}. These images are used for non-commercial academic purposes in accordance with their respective licenses. 

\paragraph{Image License Details.} Detailed information on the licensing of each image used in the figures of this paper is provided below:

\vspace{5pt}
\noindent
\Cref{fig:fig1-quali} (left)
\begin{itemize}[label=\textbullet, left=0pt, labelsep=10pt, itemsep=0pt, parsep=0pt, partopsep=0pt, align=left, topsep=0pt]
\item Title: Texting
\item Photographer: Ktoine
\item Source URL: \url{https://www.flickr.com/photos/49170045@N07/5149301465}
\item Copyright information: Antoine K
\item Creative Common license: CC BY-SA
\end{itemize}

\vspace{5pt}
\noindent
\Cref{fig:fig1-quali} (second from the left)
\begin{itemize}[label=\textbullet, left=0pt, labelsep=10pt, itemsep=0pt, parsep=0pt, partopsep=0pt, align=left, topsep=0pt]
\item Title: dk\&fish
\item Photographer: captain\_ambiance
\item Source URL: \url{https://www.flickr.com/photos/87855339@N00/2500267824}
\item Copyright information: N/A
\item Creative Common license: CC BY-NC-SA
\end{itemize}

\vspace{5pt}
\noindent
\Cref{fig:fig1-quali} (second from the right)
\begin{itemize}[label=\textbullet, left=0pt, labelsep=10pt, itemsep=0pt, parsep=0pt, partopsep=0pt, align=left, topsep=0pt]
\item Title: Frisbee 1
\item Photographer: laikolosse
\item Source URL: \url{https://www.flickr.com/photos/64952097@N00/2435927296}
\item Copyright information: N/A
\item Creative Common license: CC BY-NC
\end{itemize}

\vspace{5pt}
\noindent
\Cref{fig:fig1-quali} (right)
\begin{itemize}[label=\textbullet, left=0pt, labelsep=10pt, itemsep=0pt, parsep=0pt, partopsep=0pt, align=left, topsep=0pt]
\item Title: East Coast Marines participate in advanced motorcycle class
\item Photographer: CherryPoint
\item Source URL: \url{https://www.flickr.com/photos/71948543@N03/7789686222}
\item Copyright information: N/A
\item Creative Common license: CC BY-NC-SA
\end{itemize}

\vspace{5pt}
\noindent
\Cref{fig:method} 
\begin{itemize}[label=\textbullet, left=0pt, labelsep=10pt, itemsep=0pt, parsep=0pt, partopsep=0pt, align=left, topsep=0pt]
\item Title: I get to wear a tie today!
\item Photographer: mr.rcollins
\item Source URL: \url{https://www.flickr.com/photos/15076398@N00/6025569708}
\item Copyright information: N/A
\item Creative Common license: CC BY
\end{itemize}

\vspace{5pt}
\noindent
\Cref{fig:inp-test} (top)
\begin{itemize}[label=\textbullet, left=0pt, labelsep=10pt, itemsep=0pt, parsep=0pt, partopsep=0pt, align=left, topsep=0pt]
\item Title: Penny farthing
\item Photographer: ksuyin
\item Source URL: \url{https://www.flickr.com/photos/96256161@N00/3910801203}
\item Copyright information: Su Yin Khoo
\item Creative Common license: CC BY-NC-SA
\end{itemize}

\vspace{5pt}
\noindent
\Cref{fig:inp-test} (bottom)
\begin{itemize}[label=\textbullet, left=0pt, labelsep=10pt, itemsep=0pt, parsep=0pt, partopsep=0pt, align=left, topsep=0pt]
\item Title: DSC\_0416
\item Photographer: driver Photographer
\item Source URL: \url{https://www.flickr.com/photos/72334647@N03/7209442826}
\item Copyright information: N/A
\item Creative Common license: CC BY-SA
\end{itemize}

\vspace{5pt}
\noindent
\Cref{fig:filter-quali}
\begin{itemize}[label=\textbullet, left=0pt, labelsep=10pt, itemsep=0pt, parsep=0pt, partopsep=0pt, align=left, topsep=0pt]
\item Title: LANCE JAMES AND DOGS
\item Photographer: summonedbyfells
\item Source URL: \url{https://www.flickr.com/photos/8521690@N02/6452452909}
\item Copyright information: N/A
\item Creative Common license: CC BY
\end{itemize}

\vspace{5pt}
\noindent
\Cref{fig:gender} (left)
\begin{itemize}[label=\textbullet, left=0pt, labelsep=10pt, itemsep=0pt, parsep=0pt, partopsep=0pt, align=left, topsep=0pt]
\item Title: Lou zers
\item Photographer: orijinal
\item Source URL: \url{https://www.flickr.com/photos/48600098077@N01/4342901300}
\item Copyright information: Jaysin Trevino
\item Creative Common license: CC BY
\end{itemize}

\vspace{5pt}
\noindent
\Cref{fig:gender} (second from the left)
\begin{itemize}[label=\textbullet, left=0pt, labelsep=10pt, itemsep=0pt, parsep=0pt, partopsep=0pt, align=left, topsep=0pt]
\item Title: 091022-F-7797P-006
\item Photographer: Offutt Air Force Base
\item Source URL: \url{https://www.flickr.com/photos/61438875@N03/6427400981}
\item Copyright information: N/A
\item Creative Common license: CC BY
\end{itemize}

\vspace{5pt}
\noindent
\Cref{fig:gender} (second from the right) 
\begin{itemize}[label=\textbullet, left=0pt, labelsep=10pt, itemsep=0pt, parsep=0pt, partopsep=0pt, align=left, topsep=0pt]
\item Title: Gael Monfils at the Legg Mason Tennis Classic
\item Photographer: Kyle T.
\item Source URL: \url{https://www.flickr.com/photos/87605170@N00/6019705477}
\item Copyright information: N/A
\item Creative Common license: CC BY
\end{itemize}

\vspace{5pt}
\noindent
\Cref{fig:gender} (right)
\begin{itemize}[label=\textbullet, left=0pt, labelsep=10pt, itemsep=0pt, parsep=0pt, partopsep=0pt, align=left, topsep=0pt]
\item Title: 022
\item Photographer: tiraslee
\item Source URL: \url{https://www.flickr.com/photos/69153271@N00/3482486652}
\item Copyright information: N/A
\item Creative Common license: CC BY
\end{itemize}

\vspace{5pt}
\noindent
\Cref{fig:skin} (left)
\begin{itemize}[label=\textbullet, left=0pt, labelsep=10pt, itemsep=0pt, parsep=0pt, partopsep=0pt, align=left, topsep=0pt]
\item Title: Bear \& I @ San Diego Zoo
\item Photographer: tammylo
\item Source URL: \url{https://www.flickr.com/photos/13595617@N00/8027273225}
\item Copyright information: Tammy Lo
\item Creative Common license: CC BY
\end{itemize}

\vspace{5pt}
\noindent
\Cref{fig:skin} (second from the left)
\begin{itemize}[label=\textbullet, left=0pt, labelsep=10pt, itemsep=0pt, parsep=0pt, partopsep=0pt, align=left, topsep=0pt]
\item Title: Loop hem eruit!
\item Photographer: FaceMePLS
\item Source URL: \url{https://www.flickr.com/photos/38891071@N00/5616622936}
\item Copyright information: N/A
\item Creative Common license: CC BY
\end{itemize}

\vspace{5pt}
\noindent
\Cref{fig:skin} (second from the right) 
\begin{itemize}[label=\textbullet, left=0pt, labelsep=10pt, itemsep=0pt, parsep=0pt, partopsep=0pt, align=left, topsep=0pt]
\item Title: Jasmin 2
\item Photographer: Snake Kiddo
\item Source URL: \url{https://www.flickr.com/photos/32502055@N05/7779141980}
\item Copyright information: Daniel Lara
\item Creative Common license: CC BY
\end{itemize}

\vspace{5pt}
\noindent
\Cref{fig:skin} (right)
\begin{itemize}[label=\textbullet, left=0pt, labelsep=10pt, itemsep=0pt, parsep=0pt, partopsep=0pt, align=left, topsep=0pt]
\item Title: Raoul in the kitchen!
\item Photographer: pug freak
\item Source URL: \url{https://www.flickr.com/photos/36567879@N00/4088457159}
\item Copyright information: Donna
\item Creative Common license: CC BY-NC
\end{itemize}

\vspace{5pt}
\noindent
\Cref{fig:amt-img,fig:amt-skin,fig:amt-gender}
\begin{itemize}[label=\textbullet, left=0pt, labelsep=10pt, itemsep=0pt, parsep=0pt, partopsep=0pt, align=left, topsep=0pt]
\item Title: Hmong Woman
\item Photographer: Elliot Margolies
\item Source URL: \url{https://www.flickr.com/photos/41597157@N00/8298673953}
\item Copyright information: Elliot Margolie
\item Creative Common license: CC BY-NC
\end{itemize}

\paragraph{Creative Commons Licenses.}
Links to each Creative Commons license are provided below:
\begin{itemize}[label=\textbullet, left=0pt, labelsep=10pt, itemsep=0pt, parsep=0pt, partopsep=0pt, align=left, topsep=0pt]
\item Attribution-NonCommercial-ShareAlike (CC BY-NC-SA): \url{http://creativecommons.org/licenses/by-nc-sa/2.0/}
\item Attribution-NonCommercial (CC BY-NC): \url{http://creativecommons.org/licenses/by-nc/2.0/}
\item Attribution (CC BY): \url{http://creativecommons.org/licenses/by/2.0/}
\item Attribution-ShareAlike (CC BY-SA): \url{http://creativecommons.org/licenses/by-sa/2.0/}
\end{itemize}

\end{document}